\documentclass{article}

\PassOptionsToPackage{sort,square,numbers}{natbib}
\usepackage[final]{neurips_2022}
\usepackage[utf8]{inputenc} 
\usepackage[T1]{fontenc}    
\usepackage{hyperref}       
\usepackage{url}     
\usepackage{booktabs}       
\usepackage{amsfonts}       
\usepackage{nicefrac}       
\usepackage{microtype}      
\usepackage{xcolor}         
\usepackage{enumitem} 
\usepackage{color, colortbl}
\usepackage{xcolor}
\usepackage{pifont}

\usepackage{epstopdf}
\usepackage{graphicx}
\usepackage{subcaption}
\usepackage{pgfplots}
\usepackage{pgfplotstable}
\usepackage{hyperref}
\usepackage{comment}
\usepackage{amsmath}
\usepackage{bm}
\usepackage{subcaption}
\usepackage{xcolor}

\usepackage[capitalise]{cleveref}
\usepackage{amssymb}
\usepackage{siunitx}
\usepackage{listings}
\usepackage[utf8]{inputenc}

\usepackage{booktabs}  
\usepackage{tabu}
\usepackage{array}
\usepackage{multirow}
\usepackage{diagbox}
\usepackage{amsthm}
\usepackage{wrapfig}

\usepackage{algorithm,lipsum,changepage}
\usepackage{algpseudocode} 
\usepackage{shortcuts_ln}

\usepackage{balance}

\usepackage{enumitem}
\usepackage{microtype}
\usepackage{calc}
\usepackage{ifthen}
\usepackage{hyphenat}
\usepackage{wrapfig}
\usepackage[font=small,labelfont=bf]{caption}

\usepackage{varwidth}
\newcommand{\frank}[1]{\todo[color=yellow]{Frank: #1}} 

\newcommand{\xopt}[0]{\bm{x}^*}
\newcommand{\yopt}[0]{y^*}
\newcommand{\fopt}[0]{f^*}

\newcommand{\data}[0]{\mathcal{D}}
\newcommand{\ent}[0]{\text{H}}

\title{Joint Entropy Search\\ for Maximally-Informed Bayesian Optimization}
\newcommand{\opt}[0]{optimal pair}

\newcommand{\jes}[0]{\texttt{JES}}
\newcommand{\ei}[0]{\texttt{EI}}
\newcommand{\pes}[0]{\texttt{PES}}
\newcommand{\es}[0]{\texttt{ES}}
\newcommand{\mes}[0]{\texttt{MES}}

\author{Carl Hvarfner \\ \texttt{ carl.hvarfner@cs.lth.se} \\
        Lund University
       \And
       Frank Hutter \\ \texttt{ fh@cs.uni-freiburg.de} \\
        University of Freiburg \\Bosch Center for Artificial Intelligence
       \And
       Luigi Nardi \\ \texttt{ luigi.nardi@cs.lth.se} \\
        Lund University \\ Stanford University \\ DBtune
}

\begin{document}

\maketitle

\begin{abstract}
%
Information-theoretic Bayesian optimization techniques have become popular for optimizing expensive-to-evaluate black-box functions due to their non-myopic qualities. Entropy Search and Predictive Entropy Search both consider the entropy over the optimum in the input space, while the recent Max-value Entropy Search considers the entropy over the optimal value in the output space. We propose Joint Entropy Search (JES), a novel information-theoretic acquisition function that considers an entirely new quantity, namely the entropy over the joint optimal probability density over both input and output space. To incorporate this information, we consider the reduction in entropy from conditioning on fantasized optimal input/output pairs. The resulting approach primarily relies on standard GP machinery and  removes complex approximations typically associated with information-theoretic methods. With minimal computational overhead, \jes{} shows superior decision-making, and yields state-of-the-art performance for information-theoretic approaches across a wide suite of tasks. As a light-weight approach with superior results, \jes{} provides a new go-to acquisition function for Bayesian optimization. 
\end{abstract}

\section{Introduction}\label{sec:intro}
The optimization of expensive black-box functions is a prominent task, arising across a wide range of applications. \textit{Bayesian optimization} (BO)~\cite{Mockus1978, shahriari-ieee16a} is a sample-efficient approach, and has been successfully applied to various problems, including machine learning hyperparameter optimization~\citep{snoek-nips12a,NIPS2011_86e8f7ab,hvarfner2022pibo,vsehic2021lassobench}, robotics~\citep{calandra-lion14a, berkenkamp2021safety,mayr2022skill, mayr2022learning}, hardware design~\citep{nardi2019practical,ejjeh2022hpvm2fpga}, and tuning reinforcement learning agents like AlphaGo~\citep{chen_arXiv18}. In BO, a probabilistic surrogate model is used for modeling the (unknown) objective. The selection policy employed by the BO algorithm is dictated by an acquisition function, which draws on the uncertainty of the surrogate to guide the selection of the next query.

The choice of acquisition function is significant for the success of the BO algorithm.
%
A popular line of acquisition functions takes an information-theoretic angle, and considers the \textit{expected information gain} regarding the location of the optimum that is obtained from an upcoming query. \textit{Entropy Search} (ES)~\cite{entropysearch}, \textit{Predictive Entropy Search} (\pes{})~\cite{pes} and the earlier work of IAGO~\cite{Villemonteix} select queries by maximizing this quantity. While \es{} and \pes{} are efficient in the number of queries to optimize the objective, they both require significant computational effort and complex approximations of the expected information gain, which impacts their performance and practical use~\cite{pes, wang2017maxvalue}. 

A related information-theoretic family of approaches considers the information gain on the optimal objective value~\cite{Hoffman2016OutputSpacePE, wang2017maxvalue, pmlr-v80-ru18a}. \textit{Max-value Entropy Search} (\mes{})~\cite{wang2017maxvalue} was the first information-theoretic approach to have a proven convergence rate, albeit only in a noiseless problem setting. Moreover, its consideration of a one-dimensional density over the output space as opposed to a $D$-dimensional input space and a reduction in intricate approximations yielded a computationally efficient alternative to the \es{}/\pes{} line of approaches. 
Despite its empirical success, some crucial shortcomings of \mes{} have been highlighted in recent works. Its convergence rate has been disputed~\cite{pmlr-v162-takeno22a}, and crucially, it does not differentiate between the (unobserved) maximal objective value $f^*$ and the observed noisy maximum $y_{max}$~\cite{pmlr-v119-takeno20a, moss2021gibbon, nguyen22rmes, pmlr-v162-takeno22a}. As such, its assumption on the posterior distribution of the output $p(y|\data, \bm{x})$ does not hold in any setting where noise is present, and several follow-ups have been proposed to address the noisy problem setting~\cite{pmlr-v119-takeno20a, moss2021gibbon, nguyen22rmes, pmlr-v162-takeno22a}. 
 

We propose an approach which merges the \es{}/\pes{} and \mes{} lines of work, and provides an all-encompassing perspective on information gain regarding optimality. We introduce Joint Entropy Search (JES), a novel acquisition function which has the following advantages over existing infomation-theoretic approaches: 
\begin{enumerate}
    \item It utilizes two sources of information, by considering the entropy over both the optimum and the noiseless optimal value;
    \item It utilizes the full optimal observation, allowing it to rely primarily on exact computation through standard GP machinery instead of complex approximations; and 
    \item It is computationally light-weight, requiring minimal pre-computation relative to other information-theoretic approaches which consider the input space.
\end{enumerate}

Simultaneously to our work, a similar approach aimed at the multi-objective setting, was proposed by~\citet{tu2022joint}. The authors independently came up with the same JES acquisition function, with a subtly different approximation scheme to the one we present. We see our work as being complementary to theirs because we both demonstrate the effectiveness of this new acquisition function in different settings - theirs being multi-objective and batch evaluations, ours being single-objective and large levels of output noise. Our code for reproducing the experiments is available at\\ \url{https://github.com/hvarfner/JointEntropySearch}.

\section{Background and related work}

\paragraph{Bayesian optimization.}
We consider the problem of optimizing a black-box function $f$ across a set of feasible inputs $\mathcal{X}\subset\mathbb{R}^d$:
\begin{equation}
    \bm{x}^* \in \argmax_{\bm{x}\in \mathcal{X}} f(\bm{x}).
    \label{eq:boo}
    \end{equation}
We assume that $f(\bm{x})$ is expensive to evaluate, and can potentially only be observed through a noise-corrupted estimate, $y$, where $y = f(\bm{x}) + \epsilon, \epsilon \sim \mathcal{N}(0, \sigma_\epsilon^2)$ for some noise level $\sigma_\epsilon^2$. In this setting, we wish to maximize $f$ in an efficient manner, typically while adhering to a budget which sets a cap on the number of points that can be evaluated.  BO aims to globally maximize $f$ by an initial design and thereafter sequentially choosing new points $\bm{x}_n$ for some iteration $n$, creating the data $\mathcal{D}_n = \mathcal{D}_{n-1} \cup \{{(\bm{x}_n, y_n)}\}$. After each new observation, BO constructs a probabilistic surrogate model $p(f|\data_n)$ and uses that surrogate to build an acquisition function $\alpha(\bm{x}, \mathcal{D}_n)$. The combination of surrogate model and acquisition function encodes the strategy for selecting the next point $\bm{x}_{n+1}$. After the full budget of $N$ iterations is exhausted, a best configuration $\xopt_N$ is returned as either the $\argmax$ of the observed values, or the optimum as predicted by the surrogate model. 

\paragraph{Gaussian processes.} 
When constructing the surrogate, the most common choice is  \textit{Gaussian processes} (GPs)~\cite{rasmussen-book06a}. Formally, a GP is an infinite collection of random variables, such that every finite subset
of those variables follows a multivariate Gaussian distribution. The GP utilizes a covariance function $k$, which encodes a prior belief for the smoothness of $f$, and determines how previous observations influence prediction. Given observations $\data_n$ at iteration $n$, the posterior $p(f|\mathcal{D}_n)$ over the objective is characterized by the posterior mean $m_n$ and variance $s_n$ of the GP:
\begin{equation}
    m_n(\bm{x}) = \mathbf{k}_n(\bm{x})^\top(\mathbf{K}_n + \sigma_\epsilon^2\mathbf{I})^{-1}
    \mathbf{y},\quad
    s_n(\bm{x}) =  k(\bm{x}, \bm{x}) - \mathbf{k}_n(\bm{x})^\top(\mathbf{K}_n + \sigma_\epsilon^2\mathbf{I})^{-1}\mathbf{k}_n(\bm{x}),
\end{equation}
where $(\mathbf{K}_n)_{ij} = k(\bm{x}_i, \bm{x}_j)$, $\mathbf{k}_n(\bm{x}) = [k (\bm{x}, \bm{x}_1), \ldots, k(\bm{x}, \bm{x}_n)]^\top$ and $\sigma^2_\epsilon$ is the noise variance. Alternative surrogate models include random forests~\cite{smac} and Bayesian neural networks~\cite{snoek-icml15a, springenberg-nips2016}.

\paragraph{Acquisition functions.}
The \textit{acquisition function} acts on the surrogate model to quantify the attractiveness of a point in the search space. Acquisition functions employ a trade-off between exploration and exploitation, typically using a greedy heuristic to do so. Simple, computationally cheap heuristics are Expected Improvement (\ei{})~\cite{jones-jgo98a, bull-jmlr11a}. For a noiseless function, \ei{} selects the next point $\bm{x}_{n+1}$ as
\begin{equation}
    \bm{x}_{n+1} \in \argmax_{\bm{x}\in\mathcal{X}} \mathbb{E}\left[(y_n^* - y_{n+1}^*)^+\right] = \argmax_{\bm{x}\in\mathcal{X}} Zs_n(\bm{x})\Phi(Z) + s_n(\bm{x})\phi(Z),
\end{equation}
where $Z = (y_n^* - m_n(\bm{x}))/s_n(\bm{x})$. Other acquisition functions which use similar heuristics are the Upper Confidence Bound (\texttt{UCB})~\cite{Srinivas_2012}, and Probability of Improvement (PI)~\cite{pi}. A more sophisticated approach related to \ei{} is  Knowledge Gradient (KG)~\cite{frazier2018tutorial}.
\paragraph{Information-theoretic acquisition functions.}
Information-theoretic acquisition functions~\cite{entropysearch, pes, russo2014learning, wang2017maxvalue} and their various adaptations~\citep{shah2015parallel, hernandez2015predictive, belakaria2019max} seek to maximize the expected information gain $I$ from observing a subsequent query $(\bm{x}, y)$ regarding the optimum, $\xopt$. This equates to reducing the uncertainty of the density over the optimum, $p(\xopt|\data) = \mathbb{P}(\bm{x} =  \argmax_{\bm{x'}\in\mathcal{X}} f(\bm{x'})|\data)$, using the information obtained through $(\bm{x}, y)$. By quantifying uncertainty through the differential entropy $\ent$, design points are selected based on the expected reduction in this quantity over $p(\xopt|\data)$. Formally, this is expressed as the difference between the current entropy over $p(\xopt|\data)$, and the expected entropy of that density after observing the next query:
\begin{equation}\label{eq:es}
    \alpha_{\texttt{ES}}(\bm{x}) = I((\bm{x}, y); \xopt|\data) = 
 \ent[ p(\xopt|\data)]- \mathbb{E}_{y}\left[ \ent[p(\xopt|\data\cup (\bm{x}, y)] \right].
\end{equation}
By utilizing the symmetric property of the mutual information, one can arrive at an equivalent expression, where the entropy is computed with regard to the density over the output $y$, 
\begin{equation} \label{eq:pes}
      \alpha_{\text{ \pes{}}}(\bm{x}) = I(y; (\bm{x}, \xopt) |\data) = \ent[p(y|\data, \bm{x})] - \mathbb{E}_{{\xopt}}\left[ \ent[p(y|\data, \bm{x}, \xopt)] \right].
\end{equation}
Eq.~\ref{eq:es} is the original formulation used in 
ES~\cite{entropysearch} and Eq.~\ref{eq:pes} is the formulation introduced with \pes{}~\cite{pes}. Both formulations require a series of approximations and expensive computational steps to compute the entropy in the second term. For \pes{} specifically, with $n$ data points of dimension $d$, the second term is estimated through Monte Carlo (MC) methods by computing Cholesky decompositions of size $\mathcal{O}(n+d^2/2)^3$, and approximating the Hessian at the optimum for each MC sample. 

\mes{}~\cite{wang2017maxvalue} avoids this computational hurdle by considering
the information gain $I((\bm{x}, y); \yopt|\data)$ regarding the optimal value $\yopt$. As such, it computes the entropy reduction for a one-dimensional density:
\begin{equation} \label{eq:mes}
    \alpha_{\text{\mes{}}}(\bm{x}) = I(y; (\bm{x}, \yopt) |\data) = \ent[p(y|\data, \bm{x})] - \mathbb{E}_{\yopt}\left[\ent[p(y|\data, \bm{x}, \yopt)]\right]. 
\end{equation}
 Here, it is assumed that the posterior predictive distribution $p(y|\data, \bm{x}, \yopt)$ is a truncated Gaussian distribution, for which the entropy can be computed in closed form. However, $p(y|\data, \bm{x}, \yopt)$ takes this form only in a strictly noiseless setting~\cite{pmlr-v119-takeno20a, nguyen22rmes}, where it holds true that $f^* = y_{max}$, i.e. when the maximal observation and the optimal value of the objective function coincide. For noisy applications, this assumption leads to an overestimation of the entropy reduction~\cite{nguyen22rmes}. 

\section{Joint Entropy Search}\label{sec:jes}
We now present Joint Entropy Search (\jes{}), a
novel information-theoretic approach for Bayesian optimization. As for other information-theoretic acquisition functions, \jes{} considers a mutual information quantity. However, unlike its predecessors, \jes{} adds an additional piece of information: compared to \es{}/\pes{}, it adds the density over the noiseless optimal value $f^*$, and compared to \mes{}, it adds the density over $\xopt$. It utilizes a novel two-step reduction in the predictive entropy from conditioning on sampled optima and their associated values. Throughout the section, we will refer to a sampled optimum and its associated value, $(\xopt, \fopt)$, as an \textit{\opt{}}.

\subsection{Joint density over the optimum and optimal value: pictorial}
\jes{} considers the joint probability density $p(\xopt, \fopt)$ over both the optimum $\xopt$ and the true, noiseless optimal value $\fopt$. \begin{figure}[tb]
    \centering
    \includegraphics[width=\linewidth]{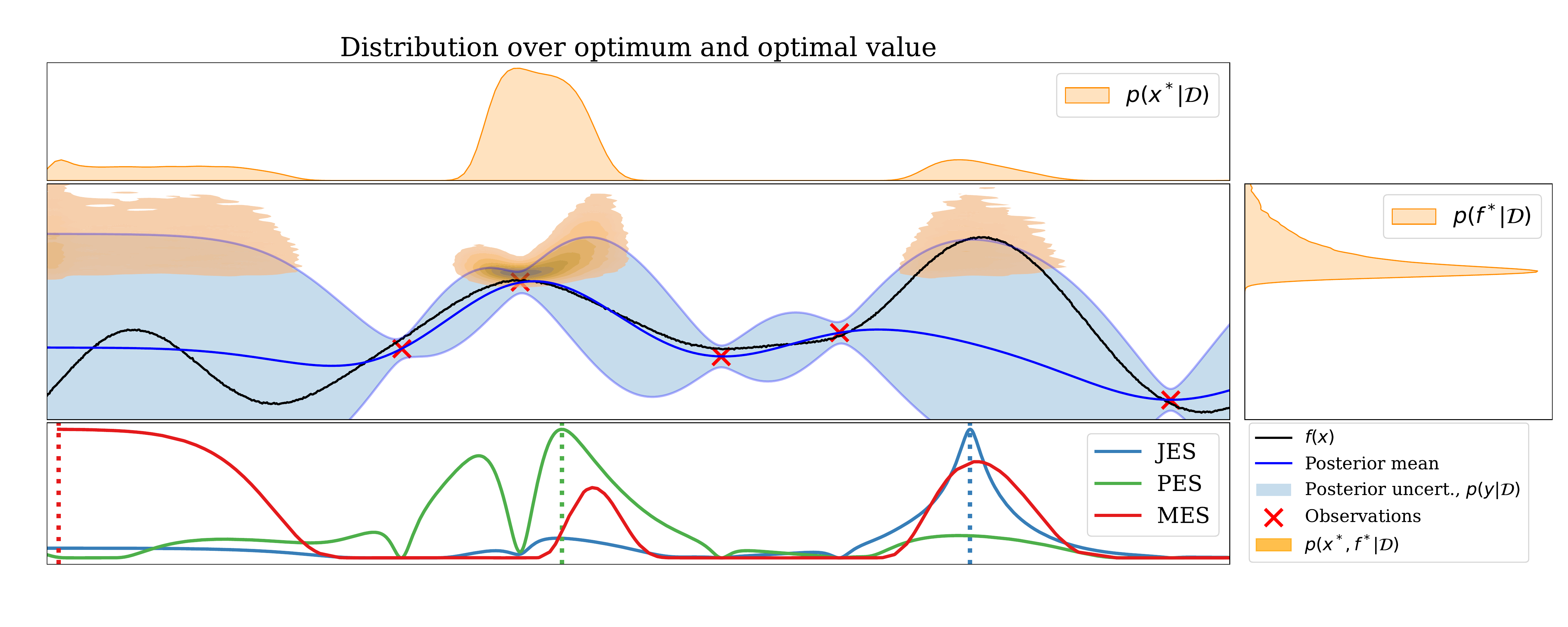}
    \vspace{-0.9cm}
    \caption{The densities considered by \es{}/\pes{} (top), \mes{} (right) and \jes{} (center) on a one-dimensional toy example. The multimodal density $p(\xopt, \fopt)$ is reduced to a heavy-tailed density over $\fopt$ for the density used by \mes{} (right), which does not capture the multi-modality of the density over the optimum. The density over $\xopt$ used by \pes{} (top) does not capture the apparent exploration/exploitation trade-off that exists between the modes. The acquisition functions and their next point selections are highlighted with dashed lines (bottom).}
    \label{fig:optdists}
    \vspace{-0.4cm}
\end{figure}
Fig.~\ref{fig:optdists} visualizes the densities $p(\xopt)$ and $p(\fopt)$, considered by \es{}/\pes{} and \mes{}, respectively, and the joint density $p(\xopt, \fopt)$, considered by \jes{}. As highlighted by the vertical dashed lines for the point selection of each strategy (bottom), \pes{} chooses strictly to reduce the uncertainty over $\xopt$, and as such, considers a region where the uncertainty over the optimal value is low. However, it can effectively determine that the right side of the local optimum is more promising to query next. \mes{} seeks to reduce the tail of the probability density over $\fopt$ (right), which in this case leads to an exploratory query. \jes{}' joint probability density over optimum and optimal value captures uncertainties over both ``where'' and ``how large'' the optimum will be. As such, it selects a point which is uncertain under both measures. As such, JES will learn about likely locations for the optimum, while simultaneously learning probable lower and upper bounds for the optimal value, which by itself yields an effective query strategy~\cite{wang2017maxvalue} and provides valuable knowledge for future queries. For the selected query in Fig.~\ref{fig:optdists}, \jes{} will learn substantially about both $\xopt$ and $\fopt$ by querying it, whereas \pes{} and \mes{} learn only about one of them. 
\frank{Hmm, actually, the query point doesn't need to be the optimum / have optimal value, but learning about it should increase our knowledge about the location and value of the optimum. Could you please reword along these lines?}

\subsection{The Joint Entropy Search acquisition function}
We consider the mutual information between the random variables $(\bm{x}^*, f^*)$ and a future query $(\bm{x},y)$: 
\begin{align}
    \label{eq:pesreform}
    \alpha_{\jes{}}(\bm{x}) &= I((\bm{x}, y); (\bm{x}^*, f^*)|\data_n)\\ 
    &= 
    \label{eq:mesreform_without_union}
    \ent[p(y|\data, \bm{x})] - \mathbb{E}_{(\xopt, \fopt)}\left[ \ent[p(y|\data, \bm{x}, \xopt, \fopt)] \right]\\ 
    &=\label{eq:mesreform}
    \ent[p(y|\data, \bm{x})] - \mathbb{E}_{(\xopt, \fopt)}\left[\ent[p(y|\data \cup{(\xopt, \fopt)}, \bm{x}, \fopt)]\right].
\end{align}

Eq.~\ref{eq:mesreform_without_union} is similar to Eq.~\ref{eq:mes} but with the addition of $\xopt$ and the replacement of $\yopt$ with $\fopt$ in the conditioning of the second term.
The expectation is computed with respect to a $D+1$-dimensional joint probability density over $\xopt$ and $\fopt$. In Eq.~\ref{eq:mesreform}, we make it explicit that the conditional density inside the expectation is obtained after 1. conditioning the GP on the previous data $\data$, plus one additional noiseless \opt{} ${(\xopt, \fopt)}$, and 2. knowing that the noiseless optimal value is in fact $\fopt$. By utilizing the complete observation ${(\xopt, \fopt)}$, we can treat it like any (noiseless) observation. As such, we quantify much of the entropy reduction by utilizing standard GP conditioning functionality. For 2., we cannot globally condition on $f(\bm{x}') \leq f^*, \forall \bm{x}'$. As such, we follow previous work~\citep{nguyen22rmes, wang2017maxvalue, moss2021gibbon, pmlr-v162-takeno22a} and enforce the condition \textit{locally} at the current query $\bm{x}$. The resulting effect is to truncate the GP's posterior over $f$ locally at $\bm{x}$, upper bounding it to $\fopt$. Notably, utilizing the fantasized observation ${(\xopt, \fopt)}$ guarantees that the conditioned optimal value $\fopt$ in \jes{} is actually obtained, rather than serving as a possibly unattained upper bound, which is typical in the MES family of acquisition functions. The expectation in Eq.~\ref{eq:mesreform} is approximated through MC by sampling $L$ \opt{}s $\{(\xopt_\ell, \fopt_\ell)\}_{\ell = 1}^L$ from $p(\xopt, \fopt)$ using an approximate version of \textit{Thompson Sampling} (\texttt{TS})~\citep{thompson1933likelihood}, as explained in Sec.~\ref{sec:sampling_optima-optimal}.  
\begin{figure}[tb]
    \centering
    \includegraphics[width=\linewidth]{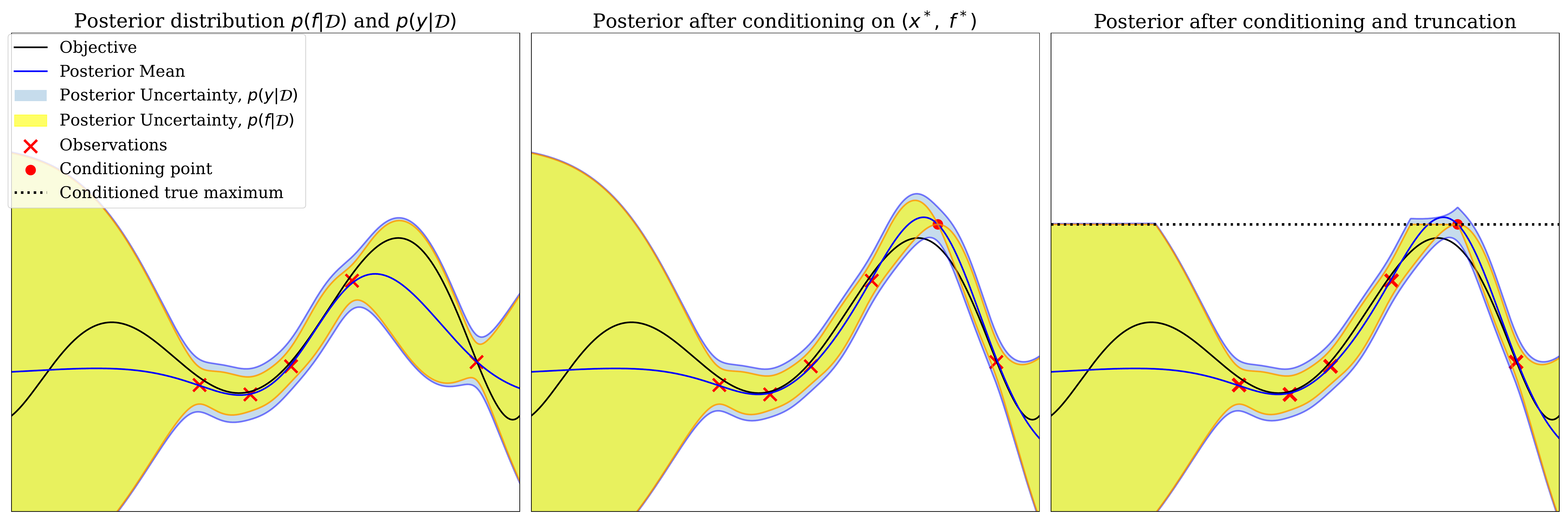}
     \caption{Step-by-step modeling when conditioning on one \opt{} $(\xopt, \fopt)$. The posterior with noise $p(y|\data)$ and without noise $p(f|\data)$ are illustrated in blue and yellow, respectively. The GP after 5 (noisy) observations, before conditioning on $(\xopt, \fopt)$ is shown on the left. In the middle panel, we draw $(\xopt, \fopt)$ and condition on it, making $p(f|\data\cup(\xopt, \fopt))$ a delta distribution at the conditioning point as the fantasized observation $\fopt$ is noiseless. Since $f^*$ is also the presumed noiseless maximum, we truncate its posterior $p(f|\data\cup(\xopt, \fopt), f^*)$ globally in the right panel. The observation noise allows for non-zero density on $p(y>f^*|\data\cup(\xopt, \fopt), f^*)$. We note that, while the noise is homoscedastic, its relative contribution to the total variance differs over the input space. As such, and since we're plotting standard deviations (not variances), the blue region is wider near observed data, where $p(f|\data)$ has lower variance.}
    \label{fig:conddists}
\end{figure}
In Fig.~\ref{fig:conddists}, the resulting posterior distribution of the two-step conditioning is shown in greater detail. As pointed out in~\cite{pmlr-v119-takeno20a, nguyen22rmes}, after conditioning on $\fopt$, the posterior predictive density over $y$ is a sum of a truncated Gaussian distribution over $f$ and the Gaussian noise $\epsilon$. The entropy reduction from the two-step conditioning yields two separate variance reduction steps over $p(y|\data, \bm{x})$: a conditioning term and a truncation term. The former is computed exactly, while the latter, generally smaller term, requires approximation, as shown in Sec.~\ref{sec:truncation}. 

\begin{wrapfigure}{r}{0.47\linewidth}

\includegraphics[width=\linewidth]{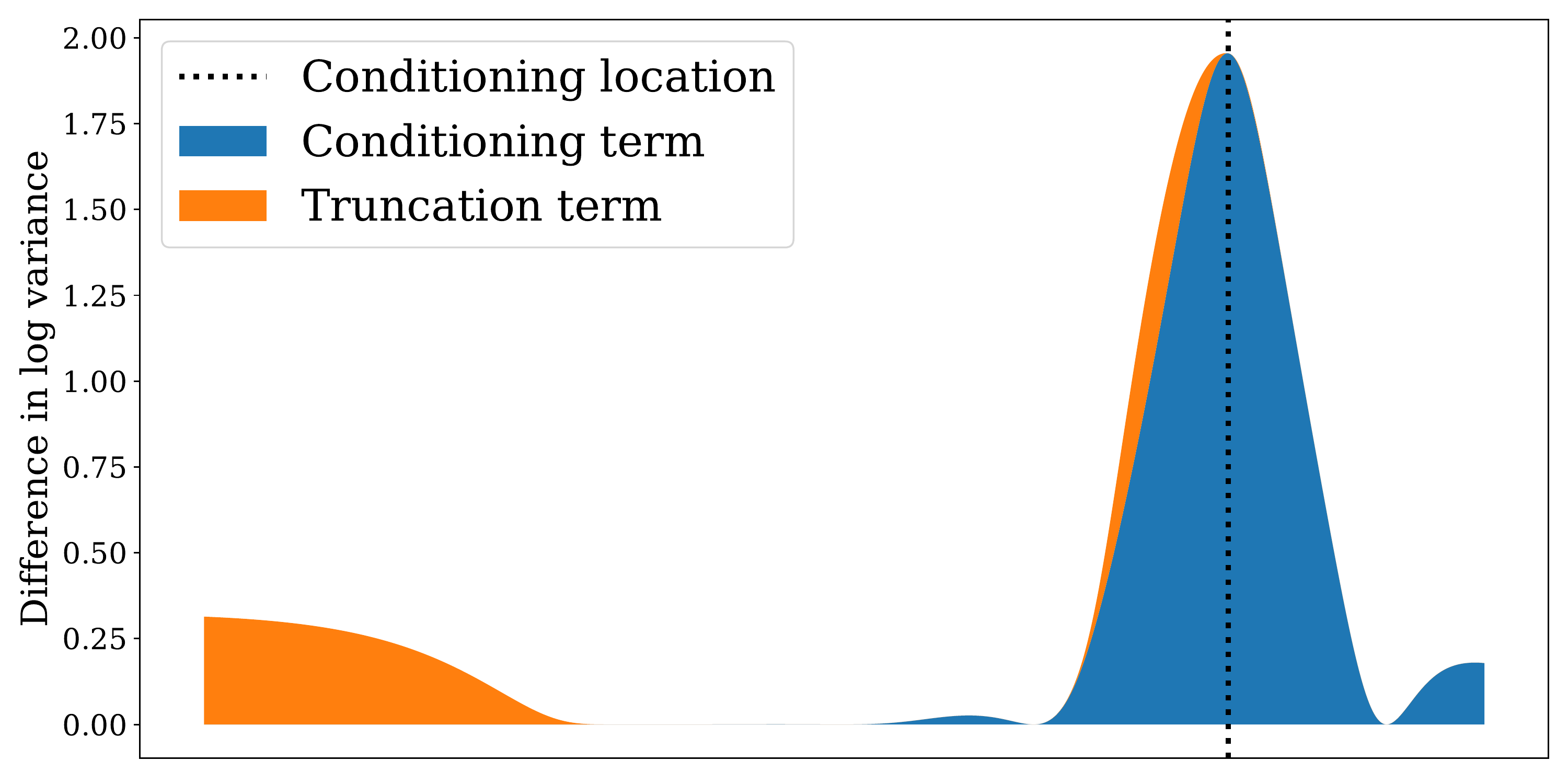}
\vspace{-0.55cm}
\caption{Reduction in log variance from the conditioning step and the truncation step as visualized in Fig.~\ref{fig:conddists}. The local conditioning term (blue), and the globally variance-reducing truncation term (orange).}
\vspace{-0.7cm}
\label{fig:entropies}
\end{wrapfigure}
Fig.~\ref{fig:entropies} shows the difference in log variance over $p(y|\data, \bm{x})$ resulting from conditioning (in blue) and truncation (in orange) for the scenario in Fig.~\ref{fig:conddists}. The overall reduction is largest close to the point of conditioning, and the truncation term mainly contributes at uncertain regions far away from the conditioned point. Moreover, the magnitude of the conditioning term will rely on the prior variance at the conditioned point, as a larger prior variance will lead to a larger reduction in entropy from conditioning. As we average over \opt{}s, many such entropy reduction terms accumulate. 

\subsection{Incorporating \opt{}s}
\label{sec:sampling_optima-optimal}
To obtain samples $(\xopt, \fopt)$, we utilize an approximate variant of \texttt{TS}~\cite{thompson1933likelihood}, originally proposed in \pes{}~\cite{pes}. We utilize Bochner's theorem~\cite{bochner1959lectures}, which, for any stationary kernel $k$, asserts the existence of its Fourier dual $s(\bm{w})$. By normalizing $s(\bm{w})$, we obtain the spectral density $p(\bm{w}) = s(\bm{w}) / \alpha$, where $\alpha$ is a normalization constant. We can then write the kernel as an expectation,
\begin{equation}
    k(\bm{x},\bm{x}') = \alpha \mathbb{E}_{\bm{w}}[e^{i\bm{w}^\intercal(\bm{x}-\bm{x}')}] = 
    2\alpha \mathbb{E}_{\bm{w, b}}[\cos(\bm{w}^\intercal\bm{x} + \bm{b})\cos(\bm{w}^\intercal\bm{x}' + \bm{b})],
\end{equation}
where $\bm{b} \sim \mathcal{U}(\bm{0}, 2\pi \bm{I})$. Following ~\citet{ali07rff}, we sample $\bm{b}$ and $\bm{w}$
to obtain an unbiased estimate of the kernel $k$. From this approximation, approximate sample paths can be drawn as a weighted sum of basis functions. This form allows for fast and dense querying of the sample paths -- the $\argmax$ and $\max$ of which is an approximate draw from $p(\xopt, \fopt)$. In \pes{}, each sample $\xopt_\ell$ along with its inverted Hessian is required for computing the acquisition function. To obtain the Hessian, each sample needs to be thoroughly optimized through gradient-based optimization. \jes{} on the other hand, only requires $(\xopt, \fopt)$. As such, it can rely on cheap, approximate optimization of these samples, e.g., by densely querying sample points on a non-uniform grid.

After obtaining a set of \opt{}s $\{(\xopt_\ell, \yopt_\ell)\}_{\ell=1}^L$, \jes{} computes the conditional entropy quantity over the output $y$. Concretely, we generate $L$ GPs, each modeling a posterior density $\{p(y|\data\cup(\xopt_\ell, \fopt_\ell), \bm{x})\}_{\ell=1}^L$ conditioned on an \opt{} and previously observed data $\data$. Since each \opt{} is drawn from the current GP hyperparameter set, we know that the current hyperparameter set is the correct one even after adding the \opt{} to the data. By consequence, \jes{} can compute the updated inverse Gram matrix, $(K+\sigma^2_\epsilon)^{-1}$, through a rank-1 update, instead of solving a linear system of equations. Utilizing the Sherman–Morrison formula~\cite{sherman1950inverse}, we obtain updated Gram matrices in $\mathcal{O}(n^2)$ for each sample, as opposed to $\mathcal{O}(n^3)$ for solving the linear system of equations. 

\subsection{Approximating the truncated entropy}\label{sec:truncation}
As highlighted in the right panel of Fig.~\ref{fig:conddists}, conditioning on $f^*$ yields a truncated normal distribution $p(f|\data \cup (\xopt, \fopt), \bm{x}, f^*)$ after having locally enforced the inequality $f(\bm{x}) \leq \fopt$. The entropy, however, is computed with regard to the density over noisy observations,  $y = f + \epsilon$, which follows an Extended Skew distribution~\citep{nguyen22rmes} and as such, does not have tractable entropy. We approximate this quantity through moment matching~\citep{moss2021gibbon} of the truncated Gaussian distribution over $f$, which yields a valid lower bound on the information gain~\citep{moss2021gibbon}. Consequently, we obtain two Gaussian densities $\hat{p}(f|\data \cup (\xopt, \fopt), \bm{x}, f^*) \sim \mathcal{N}(m_{f|\fopt}, {\sigma_{f|\fopt}^2})$ and $p(\epsilon) \sim \mathcal{N}(0, \sigma_\epsilon^2)$, where $m_{f|\fopt}$ and $\sigma_{f|\fopt}^2$ are the mean and variance of the truncated Gaussian posterior $p(f|\data \cup (\xopt, \fopt), \bm{x},  f^*)$. Due to independence between $f$ and $\epsilon$ and the linearity of Gaussian distributions, we can then compute the entropy of the approximate density $\hat{p}_y$ exactly as $\ent[\hat{p}(y|\data\cup(\xopt, \fopt), \bm{x}, \fopt)] = \log(2\pi (\sigma_\epsilon^2 + \sigma_{f|\fopt}^2))$. Moreover, the variance of the truncated Gaussian $\sigma_{f|\fopt}^2$ is computed as
\begin{equation}
 \sigma_{f|\fopt}^2(\bm{x};\data \cup (\xopt_\ell, \fopt_\ell)) = \sigma^2_T(\fopt; m_n^\ell(\bm{x}), s_n^\ell(\bm{x}))
\end{equation}
where $\sigma^2_T(\alpha; \mu, \sigma^2)$ is the variance of an upper truncated Gaussian distribution with parameters $(\mu, \sigma^2)$, truncated at $\alpha$, and $m_n^\ell(\bm{x})$ and $ s_n^\ell(\bm{x})$ are the mean and covariance functions of the GP which is conditioned on the \opt{} $(\xopt_\ell, \fopt_\ell)$. The quality of the moment matching approximation is studied in greater detail in Appendix~\ref{app:proofs}. 
\subsection{Exploitative selection to guard against model misspecification}\label{sec:misspec}
As with all information-theoretic approaches, \jes{} aims to reduce the uncertainty over the location of the optimum. With this strategy, the incentive to query the perceived optimum is often lower than for heuristic approaches, such as \ei{}. In cases where the surrogate model is misspecified, information-theoretic approaches risk reducing the entropy based on a faulty belief of the optimum, which can drastically impact their performance. As a remedy, we utilize a $\gamma$-exploit approach inspired by the parallel context of AEGIS~\cite{aegis}: with probability $\gamma$, \jes{} will query the $\argmax$ of the posterior mean to confirm its belief of the location of the optimum. If the model is misspecified, these exploitative steps enable the algorithm to reconsider its beliefs, rather than continuing to act based on faulty ones. In Appendix~\ref{app:misspec}, we show how this approach can substantially improve performance in cases of surrogate model misspecification, while having negligible impact on performance in the worst case.

\subsection{Putting it all together: The \jes{} algorithm}\label{sec:combined}
For a sampled set of size $L$, containing \opt{}s $\{(\xopt_\ell, \yopt_\ell)\}_{\ell=1}^L$ and GPs with mean and covariance functions $\{m_n^\ell(\bm{x}), s_n^\ell(\bm{x})\}_{\ell=1}^L$, the expression for the \jes{} acquisition function is
\begin{align}
\label{eq:alpha_JES}        
\alpha(\bm{x})_{\jes{}} &= \ent[p(y|\data, \bm{x})]- \mathbb{E}_{(\xopt, \fopt)}\left[\ent[p(y|\data\cup{(\xopt, \fopt)}, \bm{x}, \fopt)]\right] \\
        &\approx \log(2\pi (s_n(\bm{x}) + \sigma_\epsilon^2)) - \frac{1}{L}\sum_{\ell = 1}^L \log(2\pi (\sigma_\epsilon^2 + \sigma_{f|\fopt}^2(\bm{x};\data \cup (\xopt_\ell, \fopt_\ell))),
        \label{eq:alpha_JES:approximation}
\end{align}
The first term in \ref{eq:alpha_JES:approximation} is simply the entropy of a Gaussian that can be computed in closed form. The second term contains both the conditioning term, which is exact, and the truncation, which is approximated as described in Sec.~\ref{sec:truncation}. Algorithm~\ref{alg:jes} outlines pseudocode for \jes{} in its entirety.

\begin{algorithm}[tb] 
	\caption{\jes{} Algorithm} 
	\begin{algorithmic}[1]
	\State {\bfseries Input:} Black-box function $f$, input space $\mathcal{X}$, size $M$ of the initial design, max number of optimization iterations $N$, number of posterior MC samples $L$, fraction of exploit samples $\gamma$.
	\State {\bfseries Output:} Optimized design $\bm{x}^*$.
	
	\State $\mathcal{D}_M \leftarrow \{(\bm{x}_{i}, y_i)\}_{i=1}^M$ \;\; \Comment{initial design}
	\For{\{$n=M+1,\ldots, M+N$\}}
	    \State $m(\bm{x}), s^2(\bm{x}) \label{diff}\leftarrow$ \Call{FitGP}{$\mathcal{D}_{n-1}$}
         \If{\Call{Rand}{$0, 1$} < $\gamma$ } $\quad\bm{x}_n \leftarrow \argmax_{\bm{x}\in \mathcal{X}} m_n(\bm{x})$ \Comment{as described in Sec.~\ref{sec:misspec}}
         \Else
             \For{\{${\ell=1,\ldots, L}$\}}
             
    	    \State ${(\xopt_\ell, \yopt_\ell) \leftarrow \Call{\texttt{TS}}{f}}$ \Comment{as described in Sec.~\ref{sec:sampling_optima-optimal}}
            \State ${p(y|\data_{n-1}\cup(\xopt_\ell, \fopt_\ell), \bm{x})  \leftarrow \Call{UpdateGP}{\xopt_\ell, \fopt_\ell}}$ \Comment{as described in Sec.~\ref{sec:sampling_optima-optimal}}
        \EndFor
    	 \State $\bm{x}_n = \argmax_\mathcal{X} {\alpha_{\jes{}}(\bm{x})}$ \Comment{defined in Eq.~\ref{eq:alpha_JES}}
	 \EndIf
    \State $y_n = f(\bm{x}_n) + \epsilon, \quad \mathcal{D}_{n} \leftarrow \mathcal{D}_{n-1} \cup \{(\bm{x}_{n}, y_{n})\}$ \Comment{observe next query}
\EndFor\\
\Return $\bm{x}^* \leftarrow \argmax_{\bm{x}\in \mathcal{X}} m_n(\bm{x})$

\end{algorithmic} 
\label{alg:jes}
\end{algorithm}
\section{Experimental evaluation}
\paragraph{Benchmarks.}
We now evaluate \jes{} on a suite of diverse tasks. We consider three different types of benchmarks: samples drawn from a GP prior, commonly used synthetic test functions~\cite{pes}, and a collection of classification tasks on tabular data using an MLP, provided through HPOBench~\cite{eggensperger2021hpobench}. For the GP prior tasks, the hyperparameters are known for all methods to evaluate the effect of the acquisition function in isolation. 
Consequently, we do not use the $\gamma$-exploit approach from Sec.~\ref{sec:misspec} in this case (i.e., we set $\gamma=0$ in Algorithm \ref{alg:jes}). For the synthetic and MLP tasks, we marginalize over the GP hyperparameters, and set $\gamma=0.1$. The hyperparameters of the GP prior experiments can be found in Appendix~\ref{app:experimental}, and ablation studies on $\gamma$ in Appendix~\ref{app:ablation}.

\paragraph{Evaluation criteria.}
We use two types of evaluation criteria as in~\cite{wang2017maxvalue}: \textit{simple regret} and \textit{inference regret}. 
The simple regret 
$r_n = \max_{\bm{x}\in \mathcal{X}} f(\bm{x}) - \max_{t\in \left[1, n\right]} f(\bm{x}_t)$ measures the value of
the best queried point so far. After a query, we may infer an $\argmax$ of the function, which is chosen as $\bm{x}^*_n = \argmax_{\bm{x}\in \mathcal{X}} m_n(\bm{x})$~\cite{entropysearch,wang2017maxvalue,pes}. We denote the inference regret as $r_n = \max_{\bm{x}\in \mathcal{X}} f(\bm{x}) - f(\bm{x}^*_n)$.
Since information-theoretic approaches do not necessarily seek to query the optimum, but only to know its location, inference regret characterizes how satisfying our belief of the $\argmax$ is. Notably, this metric is non-monotonic, meaning that the best guess can worsen with time. We use this metric in the ideal model benchmarking setting, when we sample tasks from a GP with known hyperparameters. We use simple regret for the synthetic test functions, as it constitutes a metric that is more robust to surrogate model misspecification. Inference regret for these tasks can be found in Appendix~\ref{app:regrets}. For the HPOBench tasks, inference regret is unobtainable. 

\paragraph{The experimental setup.}
We compare against other state-of-the-art information-theoretic approaches: \pes{}~\cite{pes} and \mes{}~\cite{wang2017maxvalue}, as well as \ei{}~\citep{jones-jgo98a}. The acquisition functions are all run in the same framework written in MATLAB, created for the original \pes{} implementation by~\citet{pes}. All synthetic experiments were run for $50D$ iterations. In the main paper, we fix the number of MC samples for MES, PES and JES to 100 each. In Appendix~\ref{app:details_num_MC_samples} we assess the sensitivity of \jes{} to this number and quantify the computational expense. In Appendix~\ref{app:experimental}, we provide all details on our experimental setup, including the runtime tests.

\subsection{GP prior samples}
\begin{figure}[!tbp]
  \centering
  \begin{minipage}[b]{0.54\textwidth}
    \includegraphics[width=.99\linewidth]{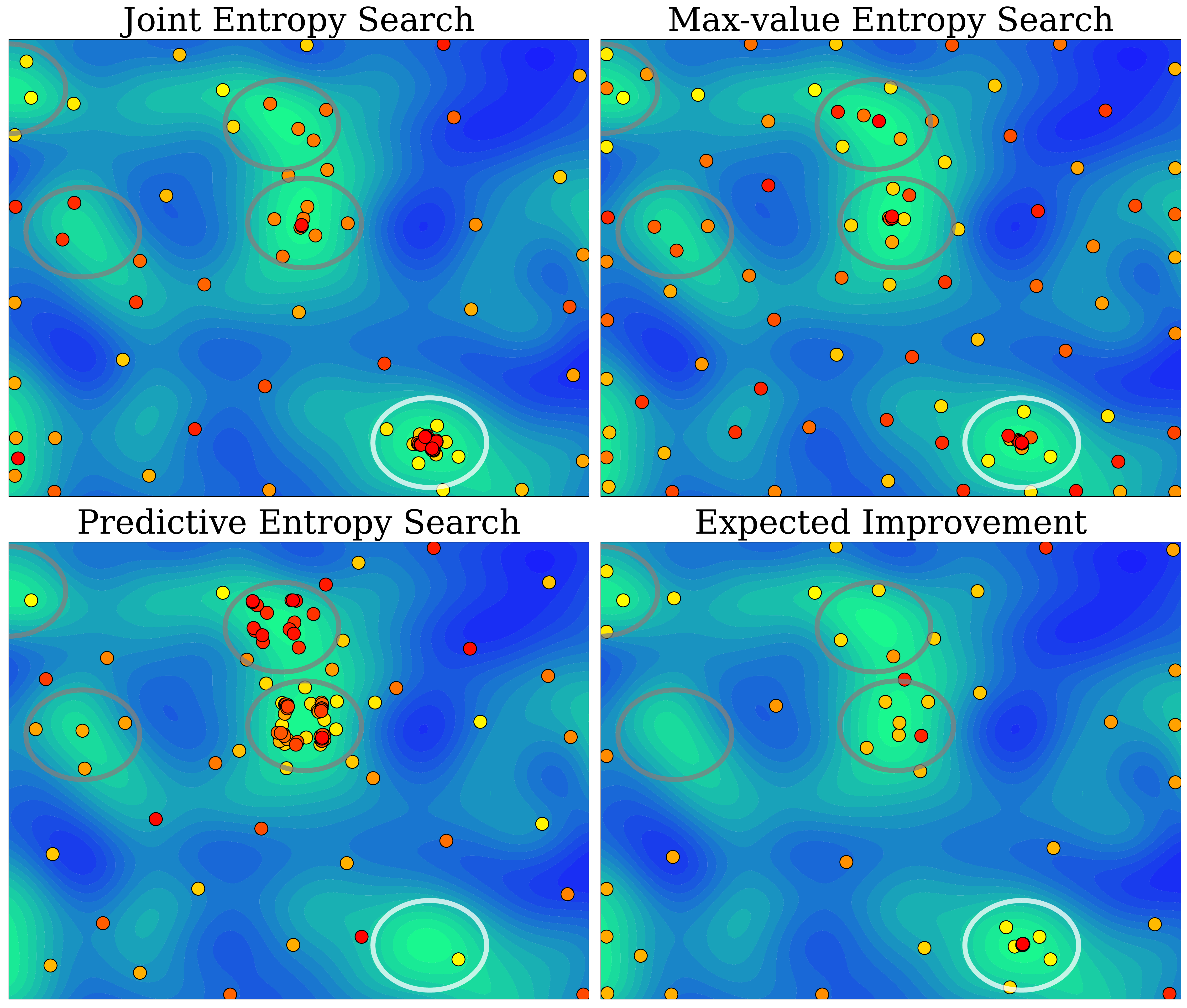}
\caption{Comparison of queries for \jes{} (top left), \mes{} (top right), \pes{} (bottom left) and \ei{} (bottom right) on a sample of a 2D GP after a 100 function evaluations. The global optimum is circled in white, and four local optima in grey. Earlier queries are colored yellow, and later queries red.}
    \label{fig:picks}
  \end{minipage}
  \hfill
  \begin{minipage}[b]{0.44\textwidth}
    \includegraphics[width=.99\linewidth]{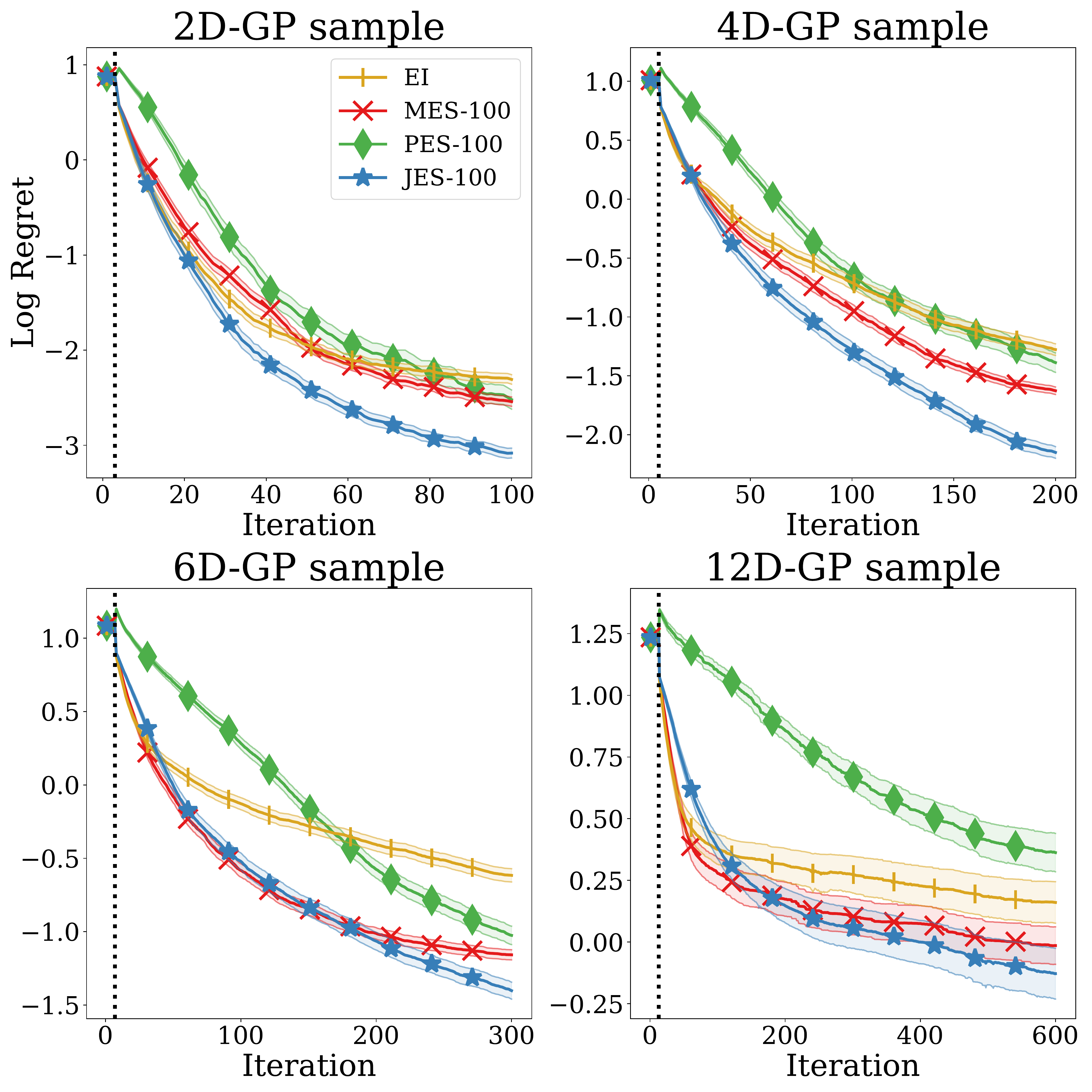}
    \caption{Comparison of \jes{}, \mes{}, \pes{} and \ei{} on  GP prior samples. We run 1000 repetitions each for 2, 4 and 6D, and 250 on 12D. Mean and 2 standard errors of log regret are displayed for each acquisition function. The vertical dashed line shows the end of the initial design phase. }
    \label{fig:gps}
  \end{minipage}
\end{figure}

We consider samples from a GP prior for four different dimensionalities: 2D, 4D, 6D, and 12D, with a noise standard deviation of $0.1$ for a range of outputs spanning roughly $[-10, 10]$. These tasks constitute an optimal setting for each algorithm, as the surrogate perfectly models the task at hand. In Fig.~\ref{fig:gps}, \jes{} demonstrates empirically the value of the additional source of information, substantially outperforming \pes{} and \mes{} on all tasks.

Fig.~\ref{fig:picks} compares \jes{} (top left) against \pes{}, \mes{} and \ei{} in terms of point selection for one repetition on a two-dimensional sample task, where all runs are initialized with $D+1$ identical random samples.
We observe that \jes{} succeeds in finding all attractive regions of the search space, and queries the region around the optimum densely, which is sensible in a noisy setting. We further notice that \ei{} (bottom right) fails to query the two circled local optima. \pes{} (bottom left) also ignores two local optima to various degrees, and tends to circle the (perceived) optimum densely, which is expensive in terms of number of evaluations. We believe this showcases a shortcoming of only considering the density over the optimum: \pes{} circles the optimum, but does not query its value. Lastly, \mes{} (top right) successfully queries all attractive regions of the space, but also samples regions that are evidently poor the most densely out of the four approaches, despite information given by earlier (brighter) samples. Since \jes{} considers the information conveyed by both \mes{} and \pes{}, it successfully excludes the apparent suboptimal regions of the space, finds all relevant optima, and queries these optima in a desirable manner. 

\begin{wrapfigure}{r}{0.5\linewidth}
\vspace{-0.45cm}
\includegraphics[width=\linewidth]{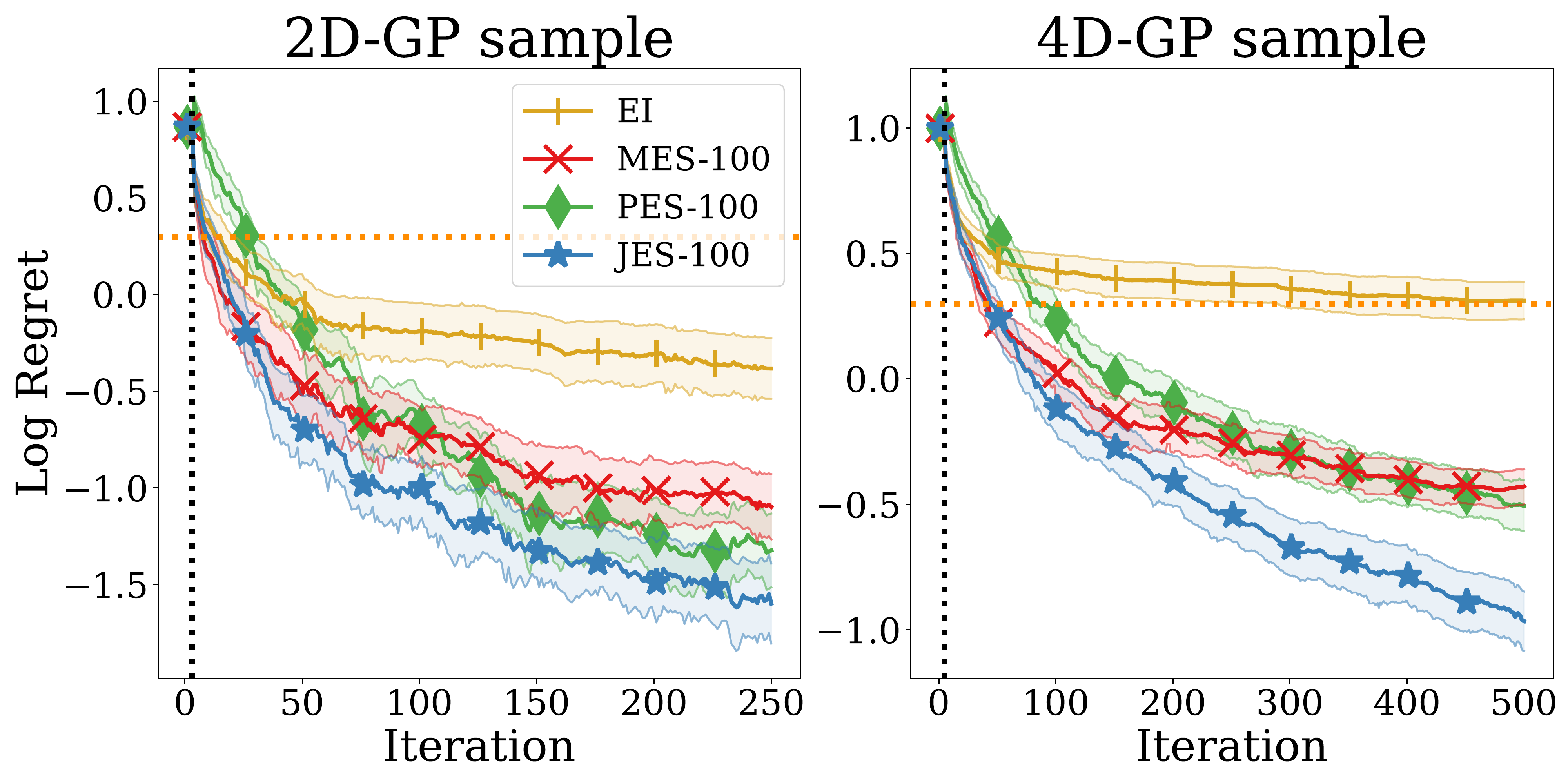}
\vspace{-0.45cm}
\caption{Evaluation of \jes{}, \mes{}, \pes{} and \ei{} on noisy ($\sigma_\epsilon^2 = 4, \text{orange})$ GP sample tasks across 100 repetitions. Mean and 2 standard errors of log regret are displayed for each acquisition function. }
\label{fig:noise}
\vspace{-1.0cm}
\end{wrapfigure}

We additionally evaluate the performance of all approaches on GP sample tasks that have a substantial amount of noise - its standard deviation roughly accounting for 10\% of the total output range. We run these tasks  with the GP hyperparameters fixed a priori for a larger number of iterations, 125D, to display the stagnation of some approaches. While \mes{} and \pes{} slow down approximately at the halfway point for both tasks, \jes{} steadily improves for the entire length of the run. This robustness to large noise magnitudes highlights the importance of intrinsically handling noisy objectives in \jes{}.

\begin{table}[htbp]

    \centering
    \begin{tabular}{c|c|c|c|c}
    \hline
    Task & JES-100 & MES-100  & PES-100 & EI\\
    \hline
    $2$D & $1.40 \pm 0.32$ & $1.03 \pm 0.19$  & $17.39 \pm 4.95$ &  $0.23 \pm 0.13$ \\ 
    $4$D & $1.50 \pm 0.37$ & $1.21 \pm 0.3$  & $34.53 \pm 8.3$ &  $0.3 \pm 0.17$ \\ 
    $6$D & $1.56 \pm 0.39$ & $1.26 \pm 0.37$  & $62.92 \pm 13.54$ &  $0.35 \pm 0.2$ \\ 
    \hline
    \end{tabular}
    \caption{Runtime of \jes{}, \mes{}, \pes{} and \ei{} on GP sample tasks of varying dimensionalities. JES is only marginally slower than MES, and orders of magnitude faster than PES.}
    \label{tab:speed}
\end{table}
In Table~\ref{tab:speed}, we display the runtime of each acquisition function on these tasks when marginalizing over 10 sets hyperparameters, and sampling 10 optima per set. We time each iteration from after hyperparameters have been sampled, up until (but excluding) the query of the black-box function. Thus, acquisition function pre-computation and optimization are included. The runtime of \jes{} is only marginally slower than that of \mes{} with Gumbel sampling, while being at least an order of magnitude faster than \pes{} for all displayed dimensionalities. 
\subsection{Synthetic test functions}

\begin{figure}[htbp]
    \centering
    \includegraphics[width=\linewidth]{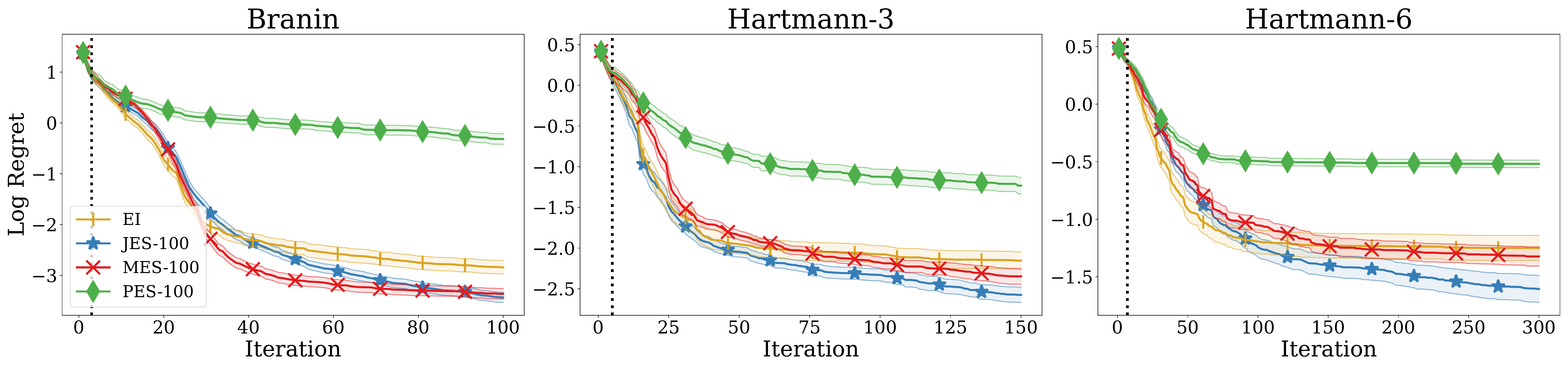}
    \caption{Comparison of \jes{}, \mes{}, \pes{} and \ei{} on Branin and Hartmann-6, $\sigma_n^2 = 0.10$. Mean and 2 standard errors of log regret are displayed for each acquisition function across 100 repetitions. The vertical dashed line represents the end of the initial design phase. }
    \label{fig:synthetic}
\end{figure}

Next, in Fig.~\ref{fig:synthetic}, we study the performance of \jes{} on three optimization test functions: Branin (2D), Hartmann (3D) and Hartmann (6D). For these tasks, we follow convention~\citep{pes, pmlr-v80-ru18a} and marginalize over GP hyperparameters. On Branin, \jes{} starts out slightly slower than MES but reaches the same performance in 100 iterations; and on the two Hartmann functions, JES performs amongst the best in the beginning and clearly best in the end. 
We note that \pes{} experienced numerical issues on Branin, and as such, we acknowledge that its performance should be better than what is reported.

\subsection{MLP tasks}
Lastly, we evaluate the performance of \jes{} on the tuning an MLP model's 4 hyperparameters for 20D iterations on six datasets. These tasks are part of the OpenML\footnote{https://www.openml.org/} library of tasks, and the HPO benchmark is provided through the HPOBench~\cite{eggensperger2021hpobench} suite. We measure the best observed classification accuracy. Notably, these tasks have a large amount of noise, which causes the performance to fluctuate substantially between repetitions. We observe that \jes{} performs substantially better on two tasks, and is approximately equal in performance to \ei{} on three, with \ei{} being superior in one task. JES displays superior or equal performance to \mes{} on all tasks, with \pes{} lagging behind.
\begin{figure}[htbp]
    \centering
    \includegraphics[width=\linewidth]{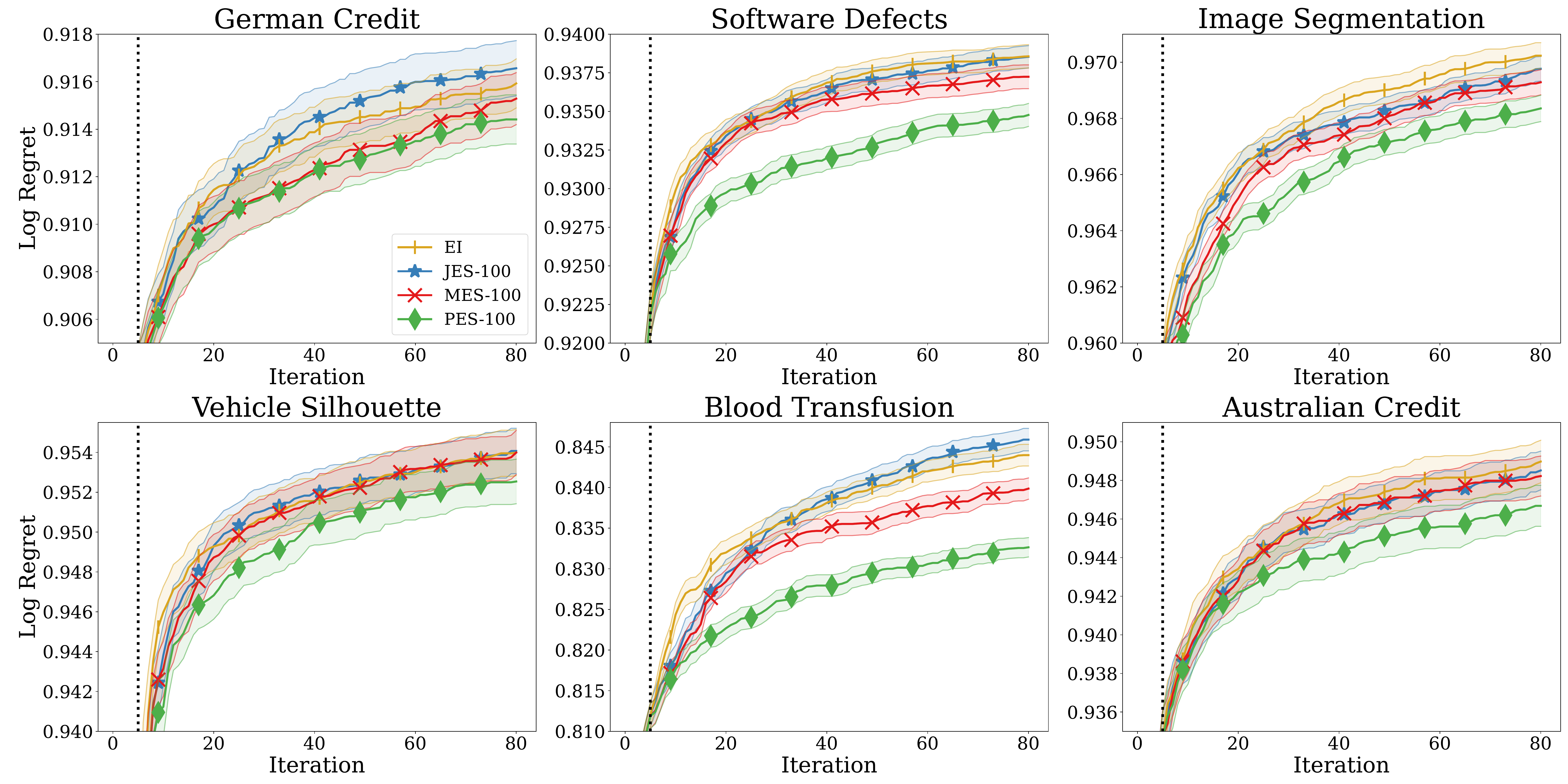}\caption{Comparison of \jes{}, \mes{}, \pes{} and \ei{} on six different MLP tuning tasks from the HPOBench suite. Mean and 1 standard error of best observed accuracy are displayed for each acquisition function across 100 repetitions. The vertical dashed line represents the end of the initial design phase.}
    \label{fig:hpo_results}
\end{figure}

\section{Conclusions}

We have presented Joint Entropy Search, an information-theoretic acquisition function that considers an entirely new quantity, namely the joint density over the optimum and optimal value. By utilizing the entropy reduction from fantasized optimal observations, \jes{} obtains a simple form for the entropy reduction regarding the joint distribution. As such, the additional information considered comes with minimal computational overhead, avoids restrictive assumptions on the objective, and yields state-of-the-art performance along with superior decision-making. We believe \jes{} to be a new go-to acquisition function for BO, and to establish a new standard for subsequent information-theoretic techniques. 

\section{Limitations and Future Work}\label{sec:limitations}
The main contribution of this paper is to provide a novel information-theoretic acquisition function which, given a sufficiently accurate model, yields impressive results. However, the non-myopic, speculative nature of information-theoretic approaches lend them to be susceptible to model misspecification, such as a poor choice of GP kernel or 
GP hyperparameters. In our view, information-theoretic approaches are possibly more susceptible to this issue than their myopic counterparts (\ei{}, \texttt{UCB}, \texttt{TS}). While we propose a remedy to stabilize and improve the acquisition function under model misspecification with the $\gamma$-exploit approach, this technique only serves to \textit{discover} misspecification and adjust accordingly, but not to inherently fix the misspecification. We believe misspecification can only be remedied by altering the surrogate model. 
It is thus very promising to combine advanced modelling techniques with information-theoretic acquisition functions, as already done with the additive GP approach utilized in conjunction with \mes{} by~\citet{wang2017maxvalue}; further promising additions would be to tackle heterogeneous noise and input warping as done by HEBO~\citep{hebo}.

We also note that, since \jes{} computes the entropy reduction from conditioning on the optimal pair, it relies on some level of noise in the objective. A surrogate model with zero noise will result in an infinite information gain for every optimal pair, which (by utilizing some random tie-breaking strategy) would make \jes{} equivalent to TS. However, if \jes{} is to be used in a completely noiseless setting, we argue that a small noise term should be added as a remedy. As this is done by default in many prominent GP frameworks~\citep{gpy2014, gardner2018gpytorch, gpstuff}, we do not view this as a major limitation of our approach. Nevertheless, improving upon this strategy would be interesting in future work.

For future work, we also envision work on the adaptation of \jes{} to various different domains, such as multi-fidelity~\citep{zhang2017information} and multi-objective optimization~\citep{belakaria2019max}, as well as the integration of user prior knowledge over the location of the optimum~\citep{hvarfner2022pibo} to accelerate optimization.

\begin{ack}
Luigi Nardi was supported in part by affiliate members and other supporters of the Stanford DAWN project — Ant Financial, Facebook, Google, Intel, Microsoft, NEC, SAP, Teradata, and VMware. Carl Hvarfner and Luigi Nardi were partially supported by the Wallenberg AI, Autonomous Systems and Software Program (WASP) funded by the Knut and Alice Wallenberg Foundation. Luigi Nardi was partially supported by the Wallenberg Launch Pad (WALP) grant Dnr 2021.0348. Frank Hutter acknowledges support through TAILOR, a project funded by the EU Horizon 2020 research and innovation programme under GA No 952215, by the Deutsche Forschungsgemeinschaft (DFG, German Research Foundation) under grant number 417962828, by the state of Baden-W\"{u}rttemberg through bwHPC and the German Research Foundation (DFG) through grant no INST 39/963-1 FUGG, and by the European Research Council (ERC) Consolidator Grant ``Deep Learning 2.0'' (grant no.\ 101045765). The computations were also enabled by resources provided by the Swedish National Infrastructure for Computing (SNIC) at LUNARC partially funded by the Swedish Research Council through grant agreement no. 2018-05973. 
Funded by the European Union. Views and opinions expressed are however those of the author(s) only and do not necessarily reflect those of the European Union or the ERC. Neither the European Union nor the ERC can be held responsible for them.
\begin{center}\includegraphics[width=0.3\textwidth]{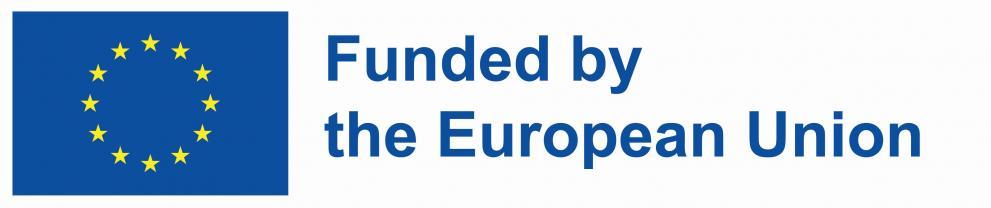}\end{center}
\end{ack}
\newpage
\bibliographystyle{abbrvnat}
\bibliography{bibliography/local,bibliography/lib,bibliography/proc,bibliography/strings}

\newpage
\section*{Checklist}


\begin{enumerate}

\item For all authors...
\begin{enumerate}
  \item Do the main claims made in the abstract and introduction accurately reflect the paper's contributions and scope?
    \answerYes{Our abstract and introduction accurately reflect our
paper.}{}
  \item Did you describe the limitations of your work?
    \answerYes{See Section~\ref{sec:limitations}}{}
  \item Did you discuss any potential negative societal impacts of your work?
    \answerYes{See Section~\ref{sec:impact}}
  \item Have you read the ethics review guidelines and ensured that your paper conforms to them?
    \answerYes{We discuss the ethics guidelines in Section~\ref{sec:impact}}{}
\end{enumerate}

\item If you are including theoretical results...
\begin{enumerate}
  \item Did you state the full set of assumptions of all theoretical results?
    \answerNA{As our method is proposed as an heuristic, we do not include theoretical results.}{}
        \item Did you include complete proofs of all theoretical results?
        \answerNA{As our method is proposed as an heuristic, we do not include theoretical results.}{}
\end{enumerate}

\item If you ran experiments...
\begin{enumerate}
  \item Did you include the code, data, and instructions needed to reproduce the main experimental results (either in the supplemental material or as a URL)?
    \answerYes{The URL to the repository where all experiments were run is included in Section~\ref{sec:intro}}{}
  \item Did you specify all the training details (e.g., data splits, hyperparameters, how they were chosen)?
    \answerYes{All experimental details are reported in Appendix~\ref{app:experimental}}{}
        \item Did you report error bars (e.g., with respect to the random seed after running experiments multiple times)?
    \answerYes{All results have error bars include.}{}
        \item Did you include the total amount of compute and the type of resources used (e.g., type of GPUs, internal cluster, or cloud provider)?
    \answerYes{See Section~\ref{sec:impact}}{}
\end{enumerate}

\item If you are using existing assets (e.g., code, data, models) or curating/releasing new assets...
\begin{enumerate}
  \item If your work uses existing assets, did you cite the creators?
    \answerYes{All the repositories that we have used or built on have been cited.}{}
  \item Did you mention the license of the assets?
    \answerYes{The licenses of the assets are mentioned in Appendix~\ref{app:experimental}}
  \item Did you include any new assets either in the supplemental material or as a URL?
    \answerYes{e include new surrogate benchmarks and frameworks in the supplementary material.}
  \item Did you discuss whether and how consent was obtained from people whose data you're using/curating?
    \answerYes{}{We did not use or release any datasets with personal data.}
  \item Did you discuss whether the data you are using/curating contains personally identifiable information or offensive content?
    \answerNA{The data we are using does not contain personal information or offensive content.}
\end{enumerate}

\item If you used crowdsourcing or conducted research with human subjects...
\begin{enumerate}
  \item Did you include the full text of instructions given to participants and screenshots, if applicable?
     \answerNA{We did not run experiments with human subjects.}
  \item Did you describe any potential participant risks, with links to Institutional Review Board (IRB) approvals, if applicable?
     \answerNA{We did not run experiments with human subjects.}
  \item Did you include the estimated hourly wage paid to participants and the total amount spent on participant compensation?
     \answerNA{We did not run experiments with human subjects.}
\end{enumerate}

\end{enumerate}

\newpage

\appendix

\title{Appendix to Joint Entropy Search\\ for Maximally-Informed Bayesian Optimization}
\maketitle
\section{Broader impact} \label{sec:impact}

Our work proposes a novel acquisition function for Bayesian optimization. The approach is foundational and does not have direct societal or ethical consequences. However, \jes{} will be used in the development of applications for a wide range of areas and thus indirectly contribute to their impacts on society. As an algorithm that can be used for HPO, \jes{} intends to cut resource expenditure associated with model training, while increasing their performance. This can help reduce the environmental footprint of machine learning research.

\section{Experimental setup}\label{app:experimental}
\paragraph{Frameworks.}
For all tasks and acquisition functions, we use the original \pes{} implementation in MATLAB by~\citet{pes}, which uses the GPStuff~\citep{gpstuff} library. The implementation optimizes the acquisition function, and the posterior mean, by sampling a dense grid of points, and uses a gradient-based optimizer to further optimize the single best point. For better accuracy, we substantially increased the number of grid points.

\paragraph{GP sample tasks.} To generate the GP sample tasks, we use a random Fourier feature~\citep{ali07rff} with weights drawn from the spectral density of a squared exponential kernel. For dimension-wise length scale $\theta_d$, output scale $\sigma^2$, and noise variance $\sigma_\epsilon^2$, the hyperparameters per task are shown in Table~\ref{tab:gps}.
\begin{table}[htbp]
    \centering
\begin{tabular}{SSSSS} \toprule
    ${D}$ & ${\theta_d}$ & ${\sigma^2}$ & ${\sigma_\epsilon^2}$ & ${\text{Approximate range}}$\\ \midrule
    2  & 0.1 & 10 & 0.01 & ${[-9, 9]}$ \\
    4  & 0.2  & 10 & 0.01 & ${[-11, 11]}$ \\
    6  & 0.3  & 10 & 0.01 & ${[-13, 13]}$ \\
    12  & 0.6  & 10 & 0.01 & ${[-18, 18]}$ \\ \bottomrule
\end{tabular}
    \caption{Hyperparameters for the generated GP sample tasks.}
    \label{tab:gps}
\end{table}
The range in the last column is a rough approximation of the magnitude of the output spanned by each GP sample. The length scales of the samples are gradually increased with each dimensionality to maintain a reasonable level of difficulty for all tasks. Since the optimal values for these tasks are unavailable, they are approximated through a dense random search, followed by local search on the most promising subset of points.

\paragraph{Runtime tests.} For the runtime tests, we run each acquisition function on 100 seeds on the 2, 4 and 6-dimensional GP sample tasks for 100 iterations each. To consider the runtime induced by each acquisition function in a realistic setting, we marginalize over 10 hyperparameter sets (including noise variance), and sample 10 optima each for \jes{}, \mes{} and \pes{}. For these experiments, the runtime of an acquisition function is considered to be the time the time from after GP hyperparameters are sampled and the GP covariance has been inverted, until the query has been selected. Thus, only acquisition function setup and acquisition function optimization are considered as part of the runtime. All acquisition functions are optimized using an identical budget of 10000 raw samples, and a subsequent gradient-based optimization around the single best point.

\paragraph{Synthetic test functions.} For the synthetic test functions, 100 sampled \opt{}s are used for each acquisition function. GP hyperparameters are marginalized over for these tasks, so an equal number of \opt{}s are sampled for each hyperparameter set. The hyperparameters are re-sampled on a fixed schedule throughout the run. Naturally, the sampled maxima were updated at each iteration. Moreover, each test function was also given a fixed amount of noise. Regret was computed not from the noisy observed value, but from the true, noiseless function value.
\begin{table}[tb]
    \centering
\begin{tabular}{SSSSS} \toprule
    {Task} & {No. hyperparameter sets} & {No. maxima per set} & {Update frequency} & {${\sigma_\epsilon^2}$}\\ \midrule
    Branin  & 20 & 5 & 5 &  0.01\\
    $\text{Hartmann (3D)}$  & 20  & 5 & 10 &  0.01\\
    $\text{Hartmann (6D)}$  & 20  & 5 & 10 &  0.01\\
    $\text{MLP classification}$  & 20  & 5 & 5 &  0.0\\ \bottomrule
\end{tabular}
    \caption{GP Hyperparameter sets and updates for the synthetic test functions and MLP tasks.}
    \label{tab:synthetic}
\end{table}

\paragraph{MLP classification tasks.} All the classification tasks have a substantial amount of noise. As noiseless objective values are unavailable, we report the observed classification accuracy. 5 hyperparameters are available in HPOBench for these tasks. However, one of them (number of layers, \textit{depth}) is held fixed due to its small integer-valued domain and the lack of integer hyperparameter support in the MATLAB framework. As seen in Tab.~\ref{tab:synthetic}, the two other integer-valued hyperparameters \textit{batch size} and \textit{width} have orders of magnitude larger domains, and can therefore reasonably be treated as continuous hyperparameters. All tasks are optimized in the [0, 1] range, and are scaled, transformed, and rounded to the nearest integer in the objective function where necessary. All non-fixed parameters are evaluated in log scale. The three tasks evaluated are \textit{Australian}, \textit{Blood-transfusion-service-center}, and \textit{Vehicle}, with HPOBench task numbers \#146818, \#10101, \#53, respectively.
\begin{table}[tb]
    \centering
\begin{tabular}{SSSS} \toprule
    {Name} & {Type} & {Range}\\ \midrule
    {Alpha (L2)}  & {Continuous} & {$[10^{-8}, 10^{-3}]$} \\ 
    {Batch size}  & {Integer} & {$[2^2, 2^8]$} \\
    {Depth}  & {Fixed to 2} & {$\{1, 2, 3\}$}\\
    {Initial learning rate} & {Continuous}  & {$[10^{-5}, 1]$}\\ 
    {Width}  & {Integer} & {$[2^4, 2^{10}]$} \\
\bottomrule
\end{tabular}
    \caption{Search space for the MLP tasks.}
    \label{tab:synthetic}
\end{table}

\paragraph{Compute resources.} All experiments are carried out on \textit{Intel Xeon Gold 6130} CPUs. Each repetition is run on a single core. In total, approximately $50,000$ core hours are used for the experiments in the main paper, and an additional $20,000$ for the appendix.

\section{Ablation studies and model misspecification}\label{app:ablation}
We provide ablation studies for the hyperparameter controlling the ratio of exploitative selections $\gamma$. Moreover, we show how $\gamma$-exploit improves inference regret by recognizing regions believed to be optimal as poor. Lastly, we display the robustness of \jes{} to the noise variance $\sigma_\epsilon^2$.

\subsection{Ablation studies}
\begin{figure}[tb]
    \centering
    \includegraphics[width=\linewidth]{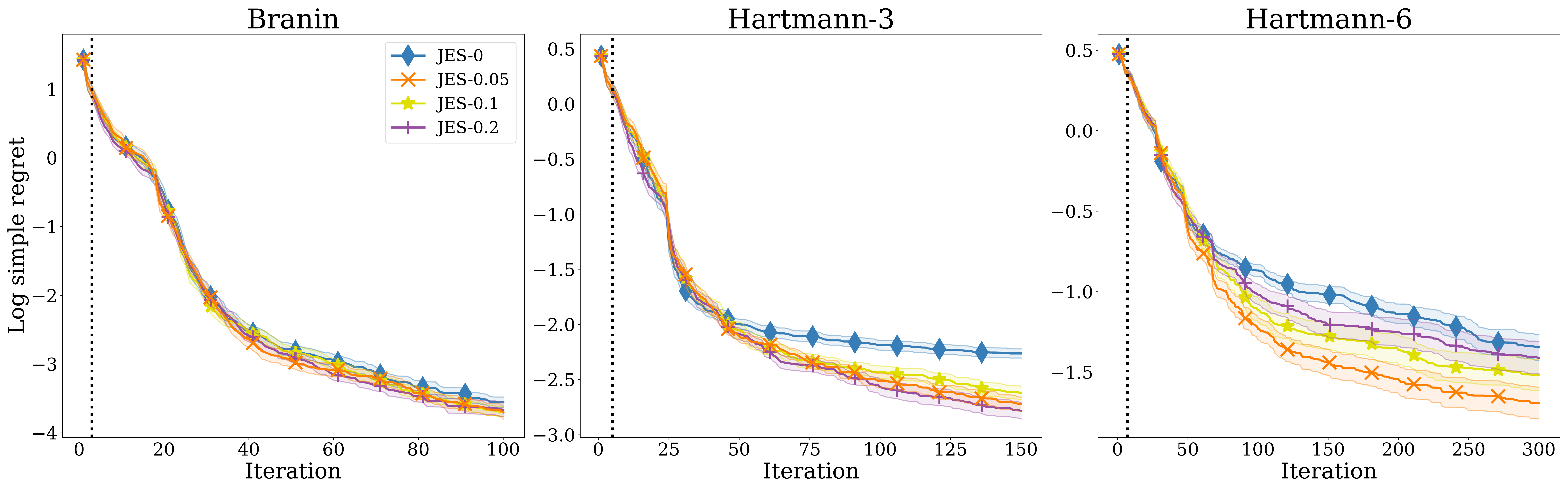}\caption{Comparison of JES with varying fraction $\gamma$ of exploitative selections. Mean and 1 standard error of log simple regret is displayed for all tasks.}
    \label{fig:gamma_simple}
\end{figure}

\begin{figure}[tb]
    \centering
    \includegraphics[width=\linewidth]{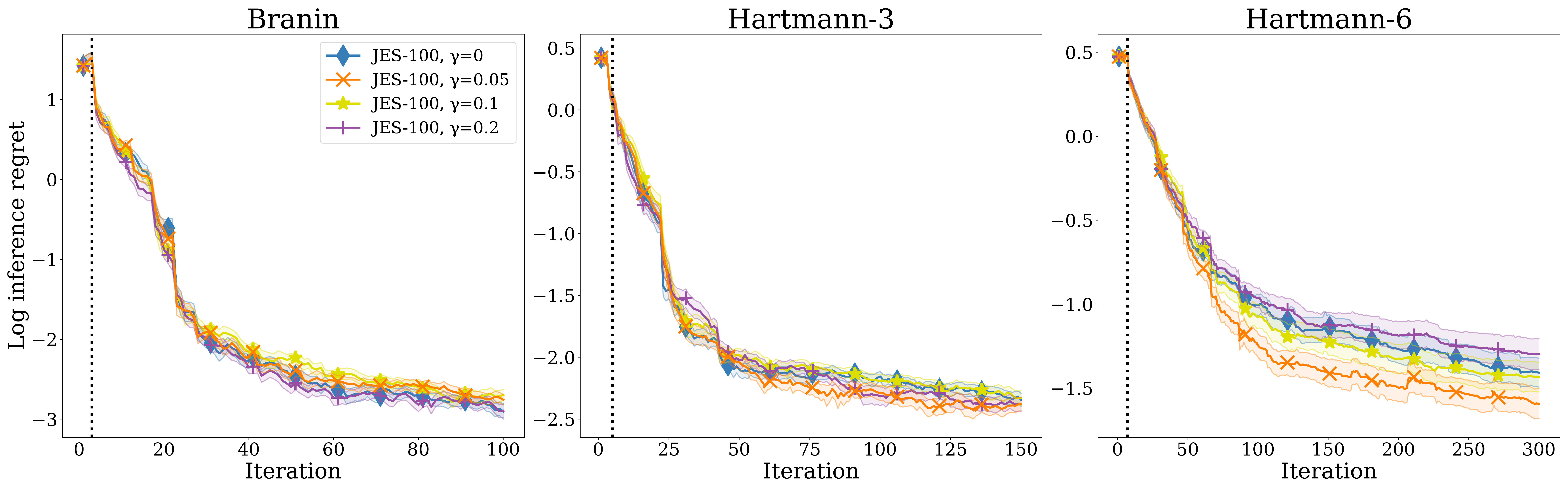}\caption{Comparison of JES with varying fraction $\gamma$ of exploitative selections. Mean and 1 standard error of log inference regret is displayed for all tasks.}
    \label{fig:gamma_inference}
\end{figure}

We provide an ablation study on $\gamma$ in terms of both the simple and inference regret of JES. While simple regret may be a more practically relevant metric, inference regret helps understand the ability of the acquisition function to successfully locate the optimum. Fig.~\ref{fig:gamma_simple} shows that $\gamma > 0 $ improves simple regret, which is to be expected from its occasional greedy selection. However, as is shown in Fig.~\ref{fig:gamma_inference}, a moderate fraction $\gamma \in \{0.05, 0.1\}$ also yields comparable or even improved inference regret on all tasks. Notably, $\gamma=0.2$ yields slightly worse performance on Hartmann (6D), but yields marginally improved performance on Branin and Hartmann (3D). As such, the $\gamma$-exploit approach not only improves performance in terms of simple regret, but yields improved inference as well. 

\subsection{Model misspecification}\label{app:misspec}
\begin{figure}[tb]
    \centering
    \includegraphics[width=\linewidth]{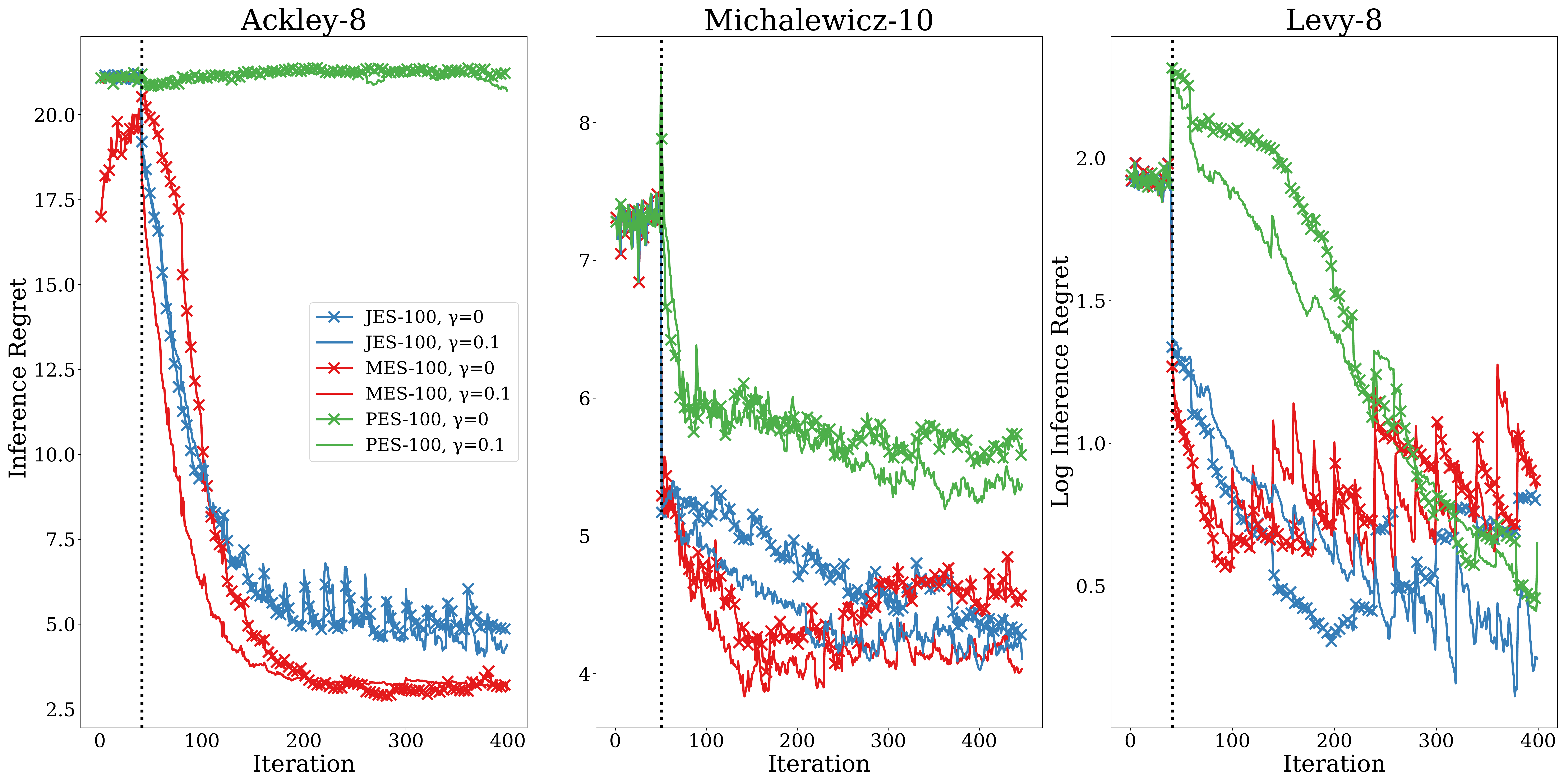}\caption{Mean of inference regret on high-dimensional synthetic functions for vanilla $\gamma$-exploit versions of \jes{}, \mes{} and \pes{}. Upward spikes signify points where the GP hyperparameters are re-sampled, and the inference regret getting worse as a result. Error bars are omitted to increase legibility.}
    \label{fig:misspec_inference}
\end{figure}

\begin{figure}[tb]
    \centering
    \includegraphics[width=\linewidth]{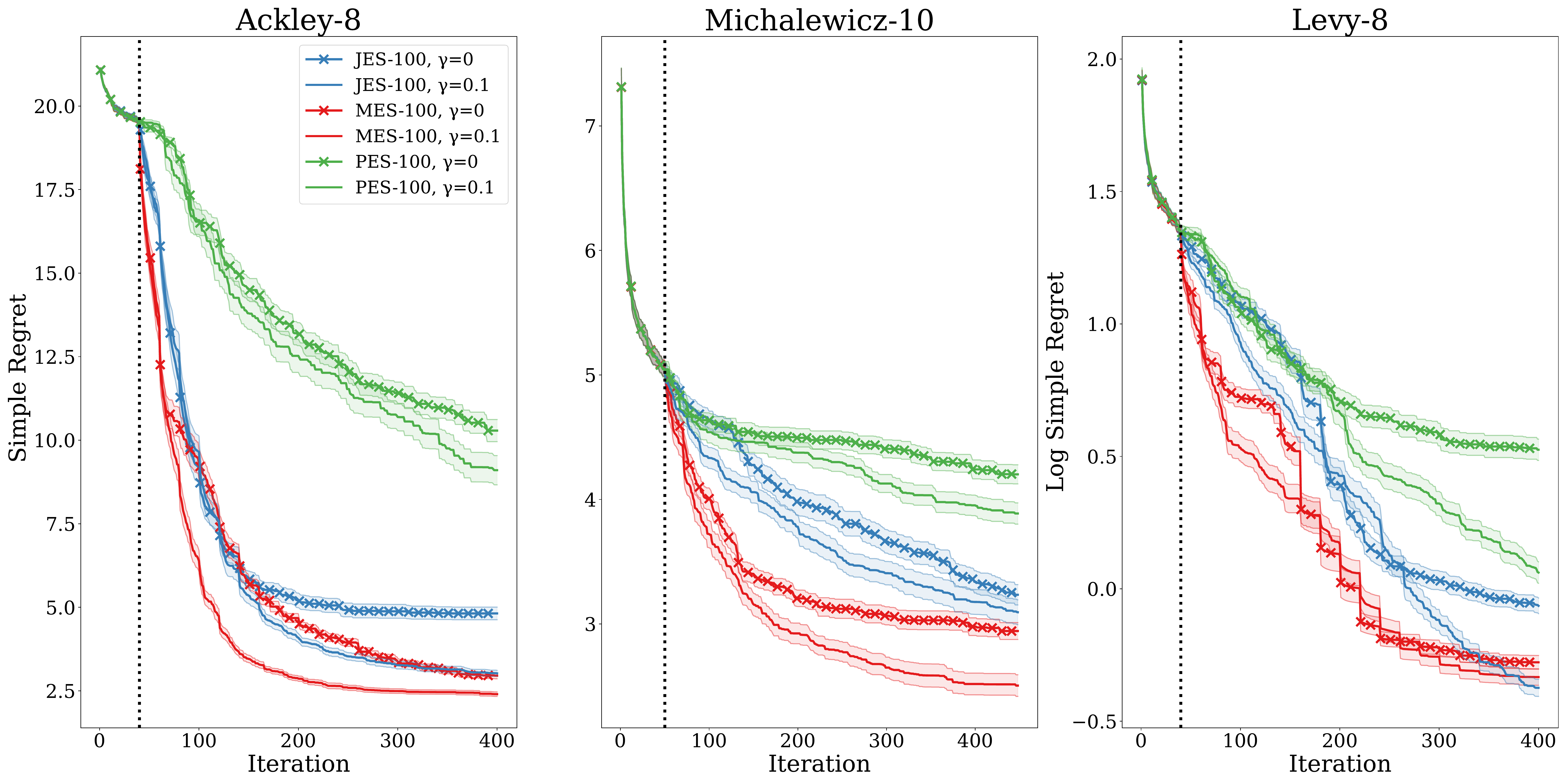}
    \caption{Mean and 1 standard error of simple regret on high-dimensional synthetic functions for vanilla $\gamma$-exploit versions of \jes{}, \mes{} and \pes{}.}
    \label{fig:misspec_simple}
\end{figure}

As mentioned in Sec.~\ref{sec:misspec}, the performance of information-theoretic methods can suffer substantially from model misspecification. In Fig.~\ref{fig:misspec_inference}, we show for \jes{}, \pes{} and \mes{} how the $\gamma$-exploit approach helps stabilize inference and improve inference regret, and yields substantially better simple regret for all methods. For Michalewicz (10D), we observe that the inference regret of \mes{} gets substantially worse after iteration 150. With the $\gamma$-exploit approach, this issue is severely reduced. In Fig.~\ref{fig:misspec_simple}, the corresponding simple regrets for \mes{} deviate at iteration 150. The same behavior can be observed for \jes{} on Levy (8D) around iteration 200. Across all test functions in Fig.~\ref{fig:misspec_inference} and Fig.~\ref{fig:misspec_simple}, an $\gamma$-exploit strategy yields comparable or improved inference regret, and strictly improved simple regret for all acquisition functions. Moreover, we observe that \mes{} is generally the top-performing acquisition function, implying that it is the most robust to model misspecification.
Using the $\gamma$-exploit approach, the acquisition function verifies whether the belief over the location of the optimum is correct under the current model hyperparameters. If it is not, it re-calibrates its belief.

\subsection{Ablation study on $\sigma_\epsilon^2$}
\begin{figure}[tb]
    \centering
    \includegraphics[width=\linewidth]{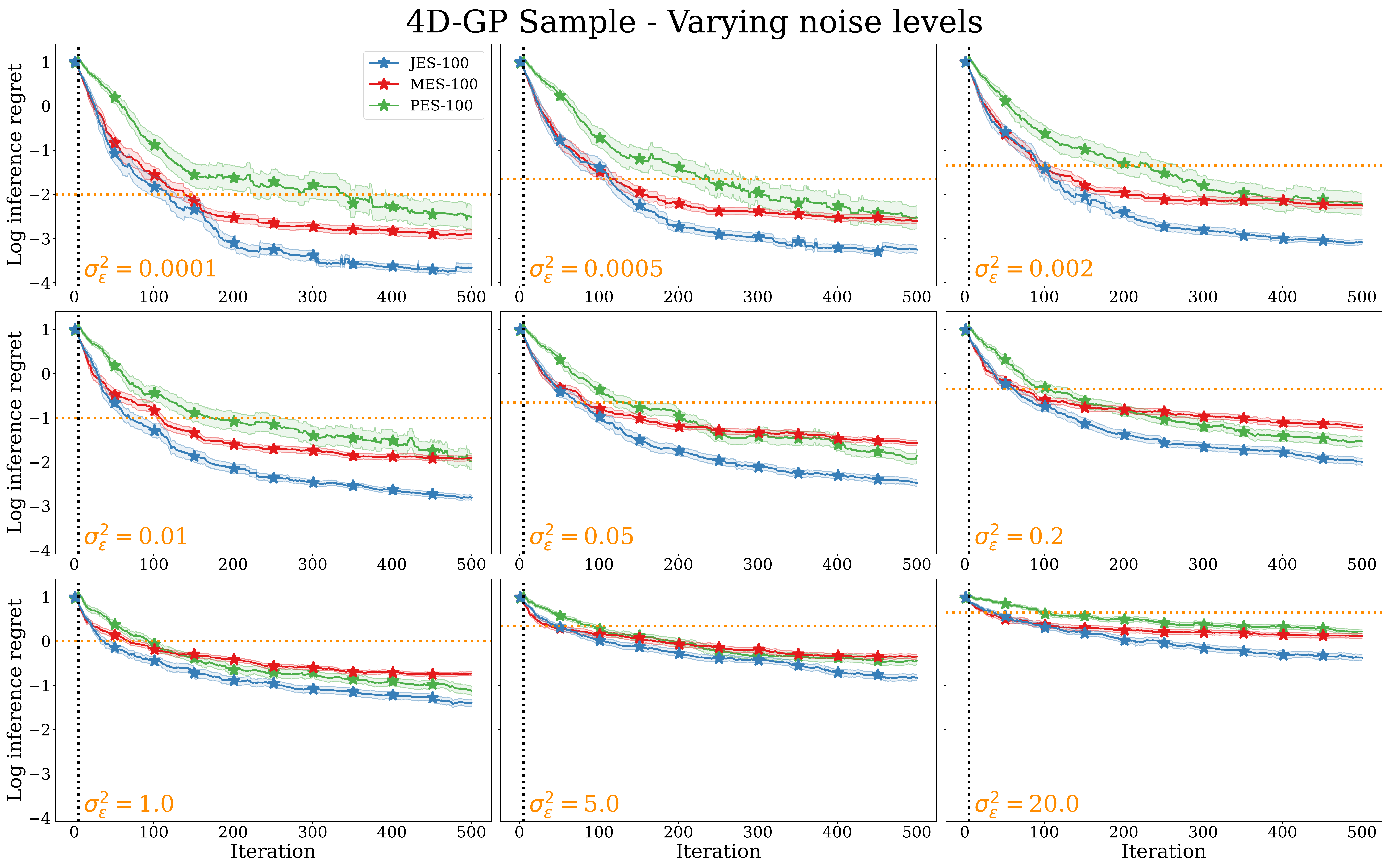}
    \caption{Evaluation of \jes{}, \mes{} and \pes{} on noisy 4D GP sample tasks across 50 repetitions for 9 different noise levels. The noise variance $\sigma_\epsilon^2$ ranges from $10^{-4}$ (top left) to $20$ (bottom right). Log noise standard deviation $\log(\sigma_\epsilon)$ is marked in dashed orange.}
    \label{fig:noiselevels4}
\end{figure}

For the noise variance ablation study, we once again consider the GP sample tasks. We fix the GP noise hyperparameter $\sigma_\epsilon^2$ to the correct value prior to the start of the experiment. In Fig.~\ref{fig:noiselevels4}, we show that the performance of \jes{} is robust with respect to the noise level, while the performance of \mes{} and \pes{} decrease more drastically as the level of noise increases.
\section{Dependence on the number of MC samples}\label{app:details_num_MC_samples}
\begin{figure}[tb]
    \centering
    \includegraphics[width=\linewidth]{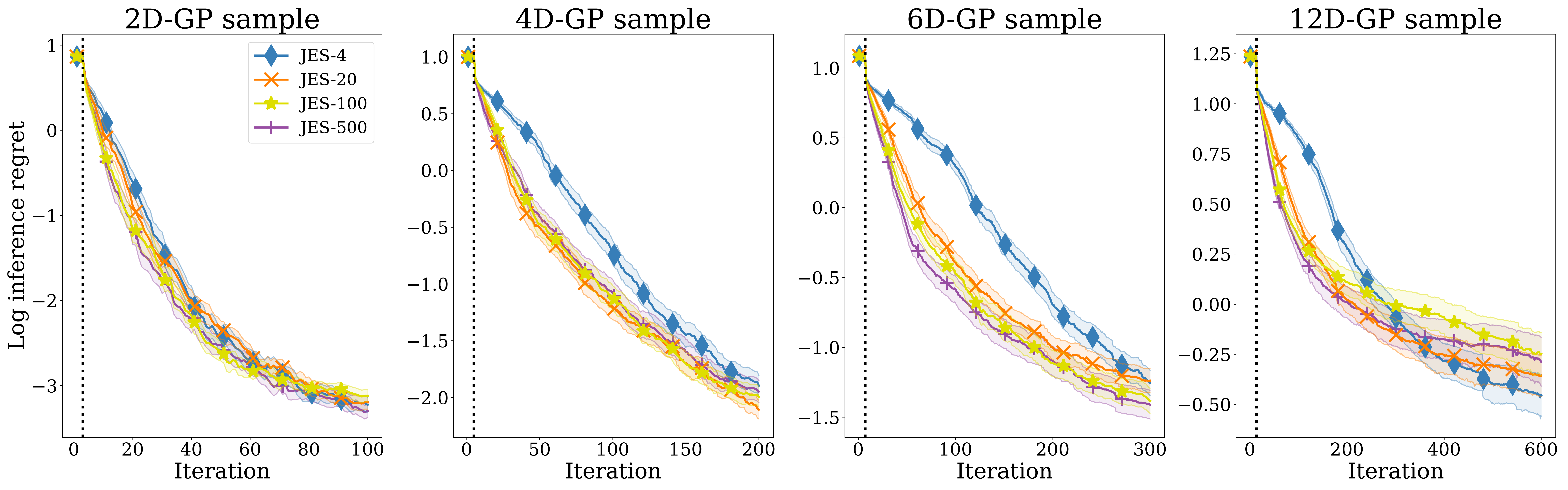}\caption{Comparison of JES with varying numbers of MC samples on GP tasks of varying dimensionalities - 10 repetitions on 10 unique tasks, for a total of 100 repetitions. Mean and 1 standard error of log inference regret is displayed for all tasks.}
    \label{fig:sample_results}
\end{figure}

We show in Fig.~\ref{fig:sample_results} the dependence of \jes{} on the number of MC samples for the GP sample tasks. \jes{} displays a slight dependence on the number of GP samples, as initial performance improves marginally for larger number of samples. This is more prominent for higher-dimensional tasks, where a lower number of samples causes a substantially slower start. This can be explained by the fact that a larger number of sampled \opt{}s are required to accurately model a higher-dimensional density. Notably, the final regret on the 12-D benchmark is better for lower numbers of samples, such as \jes{}-4. One potential explanation for this is its inability to model the joint distribution accurately. If all realized \opt{}s are close to the perceived optimum, \jes{} is almost certainly going to sample there. Moreover, since the information gain for each sample is relatively low for samples in a well-explored region, the information gain of a few samples in an unexplored region may outweight the information gain of a much larger number of samples in a well-explored region. For \jes{}-4 and \jes{}-20, it is more likely that all of the sampled \opt{}s are close to the optimum in this manner, while there is still small positive density on other parts of the search space. As such, it is possible that \jes{}-4 and \jes{}-20 over-exploit slightly in the 12-dimensional benchmark.

\section{Approximation Quality}\label{app:proofs}
As stated in Sec.~\ref{sec:truncation}, we approximate the entropy of the posterior over observations conditioned on the data and the \opt{} $\ent\left[p(y|\data\cup(\xopt, \fopt), \bm{x}, \fopt)\right]$ -- the entropy of the sum of a Gaussian and a truncated Gaussian variable -- by moment matching of the truncated distribution. We now show the quality of this approximation, and what impact it can potentially have on point selection. To do so, we utilize the results from~\citet{nguyen22rmes} regarding the density of $p(y|\data, \bm{x}, \fopt)$. We note that the approximation regards the \textit{truncation} from knowing the optimal value $\fopt$, which constitutes an additional reduction in entropy after having conditioned the GP on the new observation as in Sec.~\ref{sec:sampling_optima-optimal}. As such, the approximation considers only a fraction of the total entropy reduction, as visualized in Fig.~\ref{fig:entropies}.

\begin{figure}[tb]
    \centering
    \includegraphics[width=\linewidth]{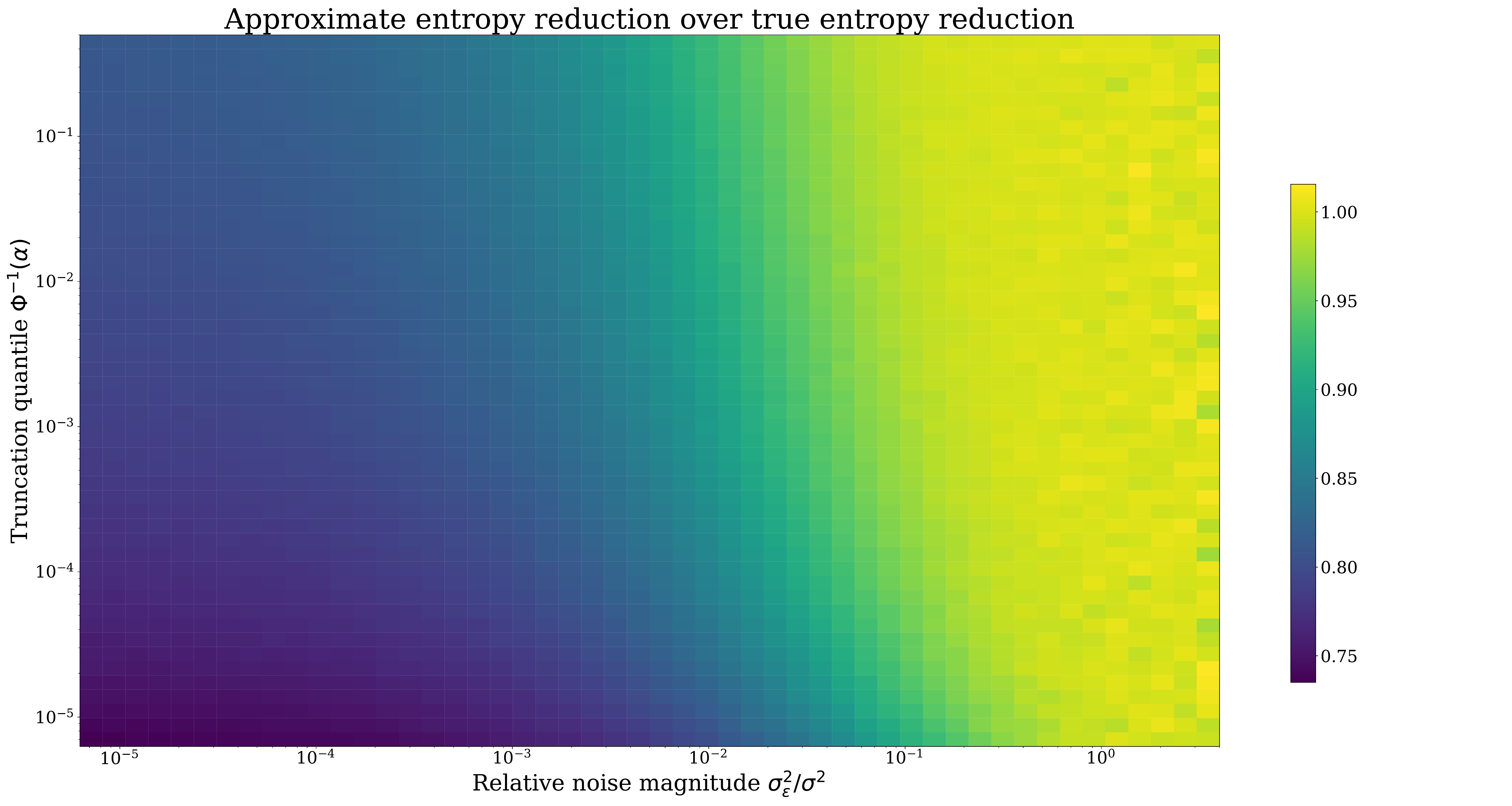}\caption{Visualization of approximation error from the moment matching approach compared to an asymptotically exact MC approach. The colormap represents the fraction of the entropy reduction resulting from truncation as approximated from moment matching divided by the entropy reduction as computed through MC. The inconsistencies in coloring in the rightmost part of the image are caused by inconsistent MC approximation.}
    \label{fig:approx_error}
\end{figure}

To establish the quality of the approximation, we compare our approach to approximating the entropy by MC. Naturally, the MC approach is more computationally expensive than moment matching, but yields an asymptotically correct result. In Fig.~\ref{fig:approx_error}, we show for varying noise levels, expressed as the noise variance ratio of the total variance, and truncation quantiles $\Phi^{-1}(\alpha)$, the difference in entropy between an MC and a moment matching approach. For example, a truncation quantile of $10^{-2}$ means that the upper $99\%$ of the density of the posterior distribution is removed as a result of truncation. 
We see that the approximate entropy reduction from moment matching is consistently lower than for the asymptotically exact MC approach. Moreover, the approximation error seemingly increases logarithmically with the truncation quantile. As such, we can expect modest underestimates of the entropy reduction when we truncate the posterior to an extreme degree, and the level of noise is low. In the right image of Fig~\ref{fig:conddists}, the posterior is severely truncated left of the conditioned location. However, the noise variance constitutes a large fraction of the total variance at this location, which means that the approximation is still accurate with respect to the true entropy reduction. The scenario represented in the bottom left corner of the grid, where we severely truncate the posterior \textit{and} the noise variance is low, does not reasonably occur in practice. The aforementioned scenario would entail sampling an \opt{} which is several orders of magnitudes \textit{worse} than the mean of the GP at uncertain (in the sense that $\sigma >> \sigma_\epsilon$) locations. Lastly, the blue region in the upper left corner of the grid represents a region where we have less truncation, and the noise is a relatively insignificant part of the total variance. Fig.~\ref{fig:approx_error} shows that the entropy reduction from truncation in this region is underestimated by approximately 15\%. As such, the approximation error leads to a slightly less explorative strategy than what a strategy with exact computation of the truncation term would provide.

\section{Regret Measures}\label{app:regrets}
We display the regret measures for the GP sample and synthetic test functions.

\subsection{GP sample tasks}
\begin{figure}[tb]
    \centering
    \includegraphics[width=\linewidth]{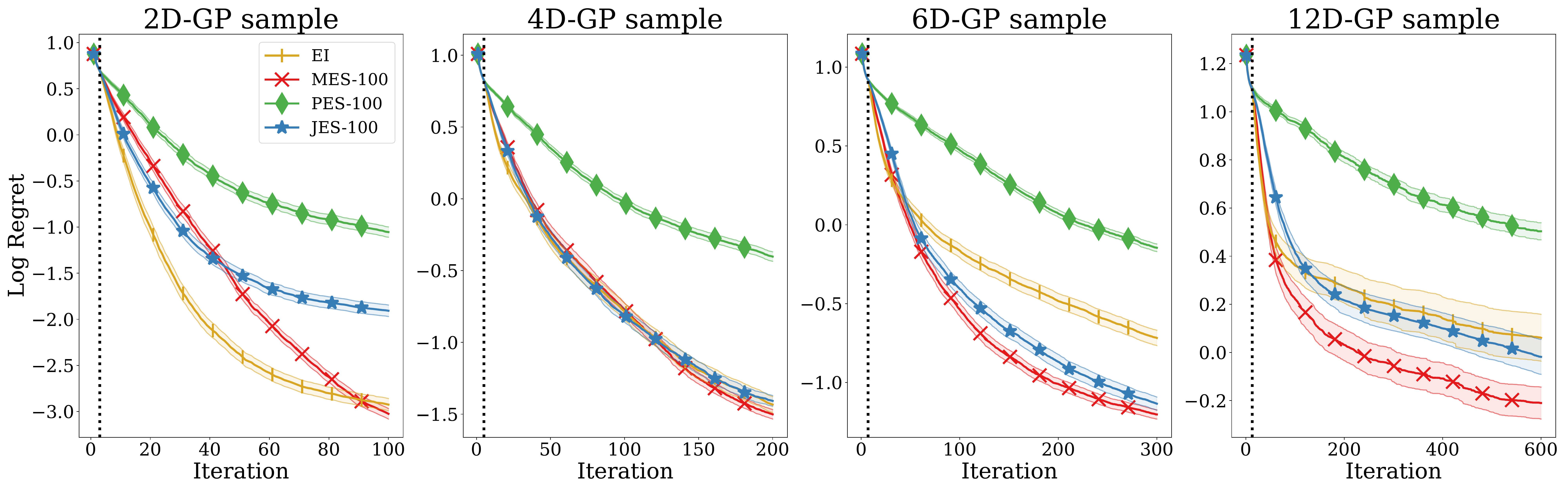}\caption{Comparison of \jes{}, \mes{}, \pes{} and \ei{} on  GP prior samples using simple regret. We run 1000 repetitions each for 2, 4 and 6D, and 250 on 12D. Mean and 2 standard errors of log regret are displayed for each acquisition function. The vertical dashed line shows the end of the initial design phase.}
    \label{fig:gp_simple_regret}
\end{figure}

In Fig.~\ref{fig:gp_simple_regret}, we show the simple regret for the GP sample tasks. We note that the simple and inference regrets for \mes{} are approximately equal, while there is a substantial difference for \jes{}. For \pes{}, the difference between simple and inference regret is the most pronounced at approximately two order of magnitude at the most.

\subsection{Synthetic test functions}
\begin{figure}[tb]
    \centering
    \includegraphics[width=\linewidth]{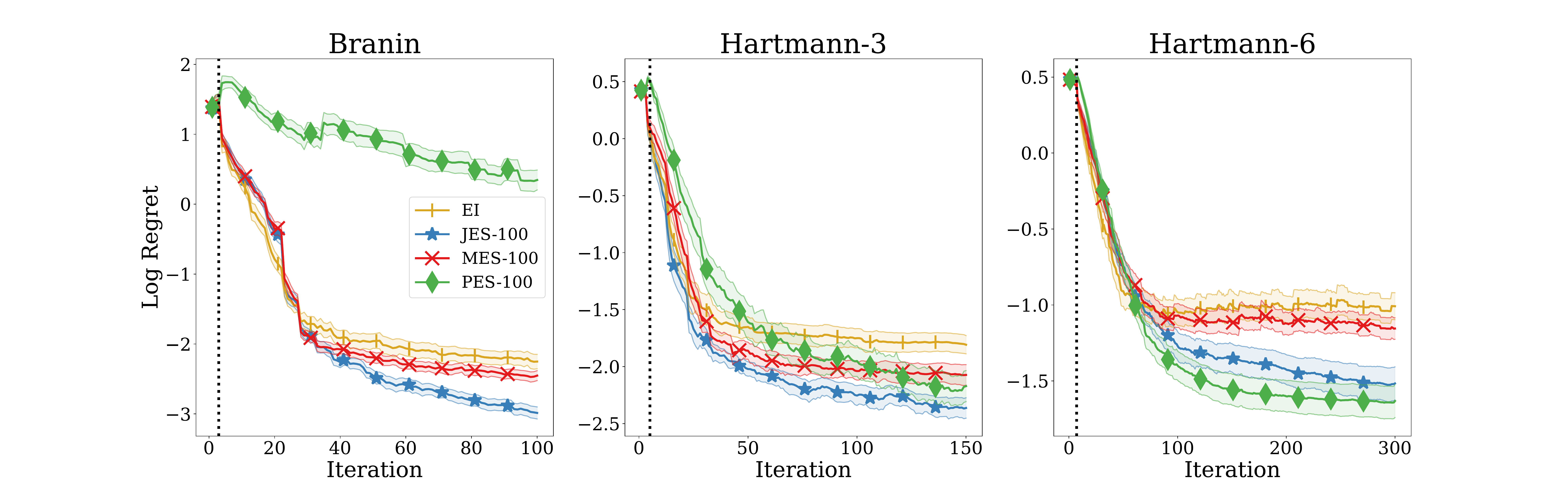}    
    \caption{Comparison of \jes{}, \mes{}, \pes{} and \ei{} on Branin and Hartmann-6, $\sigma_n^2 = 0.10$. Mean and 2 standard errors of log regret are displayed for each acquisition function across 100 repetitions. The vertical dashed line represents the end of the initial design phase. }
    \label{fig:synthetic_inference_regret}
\end{figure}

Next, we show the inference regret for the synthetic test functions in Fig.~\ref{fig:synthetic_inference_regret}. We note that he simple and inference regrets for \jes{} are approximately equal. For \pes{}, the difference between simple and inference regret is once again very pronounced. Notably, the simple regret is significantly better than the inference regret for \mes{} on Branin, implying that it does not yet have full knowledge of where the optimum is. We once again note the numerical issues of \pes{} on Branin. Moreover, the inference regret of \mes{} gets marginally worse for Hartmann (6D) from approximately iteration 100 until the end of the run, implying that its knowledge about the location of the optimum gets worse.

\end{document}